\documentclass{article}

\usepackage{arxiv}

\usepackage[utf8]{inputenc} 
\usepackage[T1]{fontenc}    
\usepackage{hyperref}       
\usepackage{url}            
\usepackage{booktabs}       
\usepackage{amsfonts}       
\usepackage{nicefrac}       
\usepackage{microtype}      
\usepackage{lipsum}
\usepackage{graphicx}
\graphicspath{{pics/}} 
\usepackage{lipsum}
\usepackage{amsmath}
\usepackage{bm}
\usepackage{amsmath}
\renewcommand{\vec}[1]{\mathbf{#1}}

\newcommand\change[1]{{{{#1}}}}

\usepackage[nolist]{acronym}

\newtheorem{theorem}{Theorem}
\usepackage{subcaption}

\begin{acronym}
    \acro{bESN}{binary ESN}
    \acro{CH}{Cayley-Hamilton}
	\acro{EoC}{Edge of Criticality}
	\acro{ESN}{Echo State Network}
	\acro{ESP}{Echo State Property}
	\acro{FIM}{Fisher Information Matrix}
	\acro{FPM}{Fractal Predicting Machine}
	\acro{LLE}{Local Lyapunov Exponent}
	\acro{LSM}{Liquid State Machine}
	\acro{MFT}{Mean Field Theory}
	\acro{MSV}{Maximum Singular Value}
	\acro{ML}{Machine Learning}
	\acro{MSO}{Multiple Superimposed Oscillator}
	\acro{NRMSE}{Normalized Root Mean Squared Error}
	\acro{PDF}{Probability Density Function}
	\acro{RC}{Reservoir Computing}
	\acro{RBN}{Random Boolean Network}
	\acro{RNN}{Recurrent Neural Network}
	\acro{RP}{Recurrency Plot}
	\acro{SI}{Supporting Information}
	\acro{SR}{Spectral Radius}
\end{acronym}

\newcommand{\matr}[1]{\bm{#1}}     

\title{Input-to-state representation \\ in linear reservoirs dynamics}

\author{
Pietro Verzelli\thanks{Referring author : pietro.verzelli@usi.ch}\\
Faculty of Informatics\\(Università della Svizzera Italiana)
\And
Cesare Alippi\\
Department of Electronics, \\Information and bioengineering
\\(Politecnico di Milano)\\
Faculty of Informatics\\(Università della Svizzera Italiana)
\And
Lorenzo Livi\\
Departments of Computer Science and Mathematics\\(University of Manitoba)\\
Department of Computer Science\\(University of Exeter)
\And
Peter Ti\v{n}o\\
School of Computer Science\\(University of Birmingham)
}

\begin{document}
\maketitle

\begin{abstract}Reservoir computing is a popular approach to design recurrent neural networks, due to its training simplicity and approximation performance. 
The recurrent part of these networks is not trained (e.g., via gradient descent), making them appealing for analytical studies {by a large} community of researchers {with backgrounds} spanning from dynamical systems to neuroscience.
However, even in the simple linear case, the working principle of these networks is not fully understood and {their design} is usually driven by heuristics. 
A novel analysis of the dynamics of such networks is proposed, which allows the investigator to express the state evolution using the controllability matrix.
Such a matrix encodes salient characteristics of the network dynamics; in particular, its rank  {represents} an input-indepedent measure of the memory {capacity} of the network.
Using the proposed approach, it is possible to compare different {reservoir} architectures and explain why a cyclic topology achieves favourable results {as verified by practitioners}.
\end{abstract}

\keywords{ Reservoir Computing \and Recurrent Neural Networks \and Dynamical Systems}

\section{Introduction}
%
%
%
%

 

Despite of being applied to a large variety of tasks, \acp{RNN} are far from being fully understood and performance improvement \change{is} usually driven by heuristics.
Understanding how the computation is conducted by the dynamics of the \acp{RNN} is an old question \cite{casey1996dynamics} which still remains unanswered, even though \change{advances were} recently achieved \cite{sussillo2013opening, ceni2018interpreting}.
The introduction of gating mechanisms (such as LSTM \cite{hochreiter1997long} and GRU \cite{cho2014learning}) dramatically improved the performance of \acp{RNN}, but the use of complex architectures makes the theoretical analysis harder \cite{tallec2018can,van2018unreasonable,jordan2019gated}.
Even for simple networks, we currently lack a sound framework to describe how the signal history is encoded in the state. 
Another relevant issue is the \emph{memory-nonlinearity trade-off} \cite{inubushi2017reservoir,dambre2012information}. 
\change{Memory capacity maximization} does not necessarily \change{imply} performance (e.g., prediction) maximization \cite{marzen2017difference}. 
In recent years, large efforts have been devoted to tackle these problems, by studying the dynamical systems behind \acp{RNN} \cite{sussillo2013opening,goudarzi2016memory,rivkind2017local,ceni2018interpreting, mastrogiuseppe2019geometrical, tino2020dynamical}.

\ac{RC} is a computational paradigm developed independently by Jaeger \cite{jaeger2001echo, jaeger2004harnessing} (\acp{ESN}) and Maas \cite{maass2002real} (\acp{LSM}) and Ti\v{n}o \cite{tino2001predicting} (\acp{FPM}).
The basic idea is to create a representation of the input signal using an untrained \ac{RNN}, called \emph{the reservoir}, and then to use a trainable \emph{readout} layer to generate the network output. 
\ac{RC} demonstrated its effectiveness in various tasks and has risen great interest in the physical computing community due to the underlying idea of \emph{natural computation}.
In particular, photonics \cite{vandoorne2014experimental} and neuromorphic computation \cite{8331848} are commonly implemented using \ac{RC}, but also a bucket of water \cite{fernando2003pattern} or road traffic \cite{ando2019road} have been used as reservoirs.
See \cite{tanaka2019recent} for a review.

The architecture simplicity makes \ac{RC} prone to theoretical investigations \cite{gonon2019risk, grigoryeva2018echo,rodan2010minimum, tino2020dynamical,goudarzi2016memory,ganguli2008memory}, which have mainly concentrated on questions about computational capabilities of whole classes of dynamical systems \cite{grigoryeva2018echo}, while little has been understood about how specific setting of the dynamical system can influence the \change{network} computational property \cite{tino2020dynamical}. 
\change{Some important theoretical results can be derived assuming that the reservoir dynamics are linear \cite{goudarzi2016memory, marzen2017difference, tino2020dynamical, ganguli2008memory, hermans2010memory}. 
This assumption can be seen as a first-order approximation of the nonlinear system, sharing some -- but not all -- of its features \cite{bollt2020explaining}.
}

In this work, we propose a novel analysis \change{ shedding light on how a linear} \ac{RNN} encodes input signals in \change{its state}.
The analysis \change{decouples the network state in two terms}: 
the \emph{controllability matrix} $\matr{\mathcal{C}}$, which only depends on the reservoir and input weights vector, and the \emph{network encoded input} $\vec{s}$, which depends on the reservoir weights and the input signal driving the system.
Writing the system in this form, allows one to decouple the properties of the reservoir topology -- encoded in $\matr{\mathcal{C}}$ -- from the specific input driving the network, encoded in $\vec{s}$. 
We analyze different reservoir topologies in terms of the nullspace of $\matr{\mathcal{C}}$ and show that the rank of $\matr{\mathcal{C}}$ is a measure for the richness of the representation of the input signal. More specifically, we show that the nullspace of $\matr{\mathcal{C}}$ is linked to the memory forgetting capacity of the network.
Based on these results, we demonstrate that a cyclic reservoir topology (forming a ring structure) is optimal. 
The claim is corroborated by empirical evidence.

\change{The remaining of the paper is organized as follows. 
In Section~\ref{sec:RC} the \ac{RC} paradigm is presented, along with the the various reservoir topologies studied in this work.
Section~\ref{sec:controllability} develops a representation of the network output in terms of its controllability matrix.
Section~\ref{sec:encoded_input} is devoted to the analysis of the input representation in the network's states, while Section~\ref{fig:nullspace} discusses the memory of the network in terms of the rank of the controllability matrix. 
In Section~\ref{sec:memory_curves} some experiments are conducted to validate are claims.
Finally, Section~\ref{sec:conclusions} draws the conclusion.
The Appendices are dedicated to derivations.
}

\section{Reservoir computing}\label{sec:RC}

\ac{RC} was developed as a tool to explain the brain working principle \cite{maass2002real} and as computational paradigm to avoid the complex and expensive training procedure of \acp{RNN} \cite{jaeger2001echo}, e.g.,  based on backpropagation through time \cite{bengio1994learning,pascanu2013difficulty}. 
\ac{RC} training \change{requires to randomly generate the weights of the recurrent layer} called \emph{reservoir}, which is tuned only at the hyper-parameter level (e.g., by searching for the best-performing spectral radius of the corresponding \change{weight matrix}).
The reservoir reads the input signal through an input layer, resulting in an untrained representation of the input in the network's state.
A \emph{readout} layer is then trained to produce the \change{system} output. 
Common tasks involve time-series prediction \cite{bianchi2015short, bianchi2017overview}, simulation of dynamical systems \cite{pathak2018model}, and time-series classification \cite{prater2017spatiotemporal}.

Let $\vec{x}_k\in \mathbb{R}^n$ be the \emph{state} of a reservoir of dimension $n$ at time $k$, $\matr{W} \in \mathbb{R}^{n \times n}$ be its \emph{reservoir connection matrix} and $\vec{w} \in \mathbb{R}^n$ its \emph{input weights} vector. 
When considering left-infinite signal, the time-index $k$ runs from  $0$ to $-\infty$ , so that the driving \emph{input signal} is $\vec{u}= (u_0, u_{-1}, u_{-2}, \dots)$, $u_{-k} \in \mathbb{R}$.

In the linear case, the reservoir evolves according to:
\begin{equation} \label{eqn:reservoir}
    \vec{x}_{k} = \matr{W}\vec{x}_{k-1} + \vec{w}u_{k}
\end{equation}

By recursively applying \eqref{eqn:reservoir} we get
\begin{align}
      \vec{x}_{0} &= \matr{W}x_{-1} + \vec{w}u_{0} 
      = \matr{W}^2 \vec{x}_{-2} + \matr{W}\vec{w}u_{-1}+ \vec{w}u_{0} \\
      &= \matr{W}^3 \vec{x}_{-3} + \matr{W}^2\vec{w}u_{-2}+ \matr{W}\vec{w}u_{-1}+ \vec{w}u_{0}    
\end{align}
i.e.,
\begin{equation} \label{eqn:res_general}
    \vec{x}_0 = \sum_{k=0}^\infty \matr{W}^k \vec{w}u_{-k}
\end{equation}{}

One usually relies on a (trained) linear readout $\vec{r} \in \mathbb{R}^n$ to generate the system output $y_{-k}$, so that at time $0$
\begin{equation} \label{eqn:readout}
    y_0 = \vec{r} \cdot \vec{x}_0 = \vec{r}  \cdot \sum_{k=0}^\infty \matr{W}^k \vec{w}u_{-k} 
\end{equation}

\begin{figure}
    \centering
    \includegraphics[width =\linewidth]{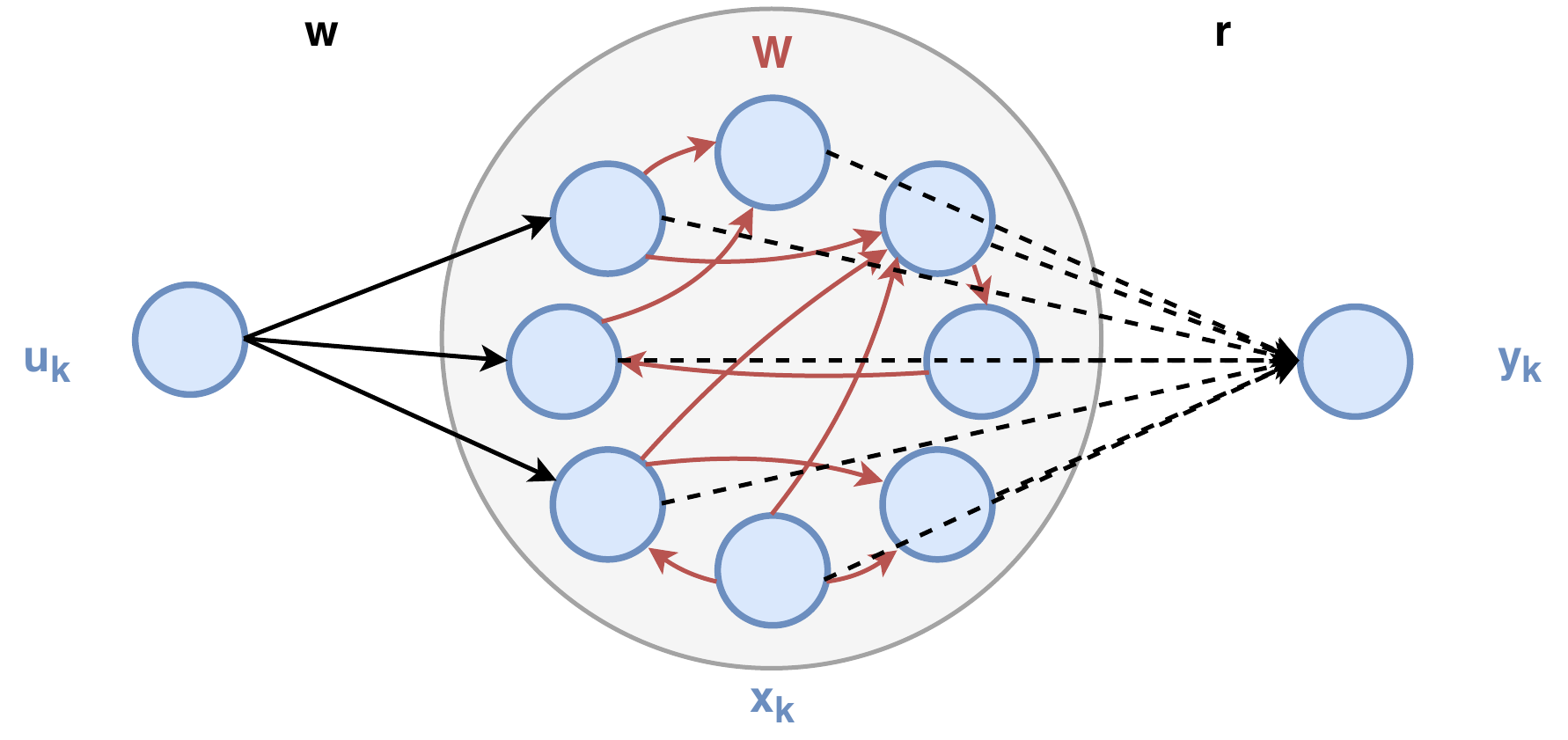}
    \caption{Schematic representation of a \ac{RC} with a \change{scalar} input, a uni-dimensional output and $N=8$ neurons in the reservoir. Fixed connections are drawn using solid lines while dashed lines represent learned connections.}
    \label{fig:RC_diagram}
\end{figure}

In the seminal paper \cite{jaeger2001echo}, training is based on a simple least-square regression. More sophisticated techniques where later introduced, including some forms of regularization \cite{reinhart2012regularization} and online-training procedures \cite{sussillo2009generating}. 
For simplicity, we have only described the case in which the input and the output are uni-dimensional, but the proposed approach can be easily generalized to the multidimensional case.
In Fig.~\ref{fig:RC_diagram}, a schematic representation of the \ac{RC} architecture is depicted.
{}

Even though a reservoir layer does not require \change{a proper training procedure}, hyper-parameters tuning must be carried out in  to improve performance.
The most studied hyper-parameter is the \ac{SR} $\rho(\matr{W})$ \cite{basterrech2017empirical, jiang2019model, caluwaerts2013spectral} which is the largest eigenvalue in magnitude of $\matr{W}$, related to the \ac{MSV} $\sigma_{\text{max}}(\matr{W})$. 
Other hyper-parameters refer to the input and output scaling factors, the sparsity degree of $\matr{W}$ and the input bias.
Moreover, the update equation \eqref{eqn:reservoir} is usually chosen to be non-linear, i.e., $ \vec{x}_{k} = \phi(\matr{W}\vec{x}_{k-1} + \vec{w}u_{k})$ and different choices of the nonlinear  transfer function  $\phi$ can be explored \cite{verzelli2019echo}. 

Many studies were devoted to understand how these hyper-parameters affect the dynamics of the network and its computational capabilities.
In particular, it appears that the hyper-parameter space can be divided into a region where the dynamics are ``regular'' (meaning that they are stable with respect to the inputs driving the system) and another one where they are ``disordered'' (meaning that they are unstable and do not provide a representation for the input) \cite{yildiz2012re}.
The narrow region separating these two regimes is known in the literature as Edge of Chaos or \ac{EoC} \cite{livi2018determination,valverde2015structural,verzelli2018characterization} and appears to be common to a large variety of complex systems beyond RNNs~\cite{langton1990computation, cocchi2017criticality, prokopenko2011relating}. 

In the following, we introduce different reservoir architectures that are commonly found in the literature.

\subsection{Delay line} \label{subsec:delay_line}
In a \emph{delay line} each neuron is connected the subsequent one to form a chain-like architecture, so that the reservoir connection matrix $\matr{W}_{\text{d}}$ reads:
        \begin{equation}
            W_{\text{d}, ij} = \delta_{i,j-1}
        \end{equation}
where $\delta$ is the \emph{Kronecker delta}. Note that the last neuron in the chain is not connected to the first one.
Moreover, the input weights vector is $\vec{w}_\text{d} = (1,0,0,\dots,0)$, meaning that the input enters the network only through the first neuron of the chain.
Mathematically, such a model setting corresponds to the $n$-th order AR model, \change{which is a really popular and studied tool in time series analysis and system identification \cite{bittanti2019model}.}

\subsection{Cyclic reservoir} \label{subsec:cycli}
A reservoir is said to be cyclic when each neuron is connected to another one, in a way that they form a ring. 
The reservoir matrix of a cyclic reservoir has the form:
\begin{equation}
    W_{\text{c},ij} = \delta_{i,j-1}
\end{equation}
where, with an abuse of notation, $\delta_{0,-1}:=\delta_{0,n-1}$.
\change{From the product of Kroenecker deltas, it follows that}:
\begin{equation} 
    W^2_{\text{c},ij} = \sum_k W_{\text{c},ik}W_{\text{c},kj}
    =  \sum_k \delta_{i,k-1}\delta_{k,j-1}
    =  \delta_{i,j-2}
\end{equation}
as well for higher powers.

\subsection{Random reservoir} \label{subsec:random}

In a random reservoir the entries of the reservoir matrix $\matr{W}_{\text{r}}$ are independent random variables.
In the sequel, we consider the generic $ij$ component of the matrix to be drawn from a Gaussian distribution. 
\begin{equation}\label{eqn:res_random}
    W_{\text{r},ij} \sim \mathcal{N}\left(0, \frac{\rho^2}{n}\right)
\end{equation}

With this choice, the  expected value for the \ac{SR} $ \langle \rho(\matr{W}_\text{r}) \rangle= \rho$ \cite{rivkind2017local} and \ac{MSV} $\langle \sigma_{\text{max}}(\matr{W}_\text{r}) \rangle= 2 \rho$ \cite{rudelson2010non} .

\subsection{Wigner reservoir}\label{subsec:wigner}
The diagonal elements are distributed as in \eqref{eqn:res_random}, i.e., $W_{\text{w},ii} \sim \mathcal{N}(0, \rho_1^2/n)$, while off-diagonal elements follow:
\begin{equation}
    W_{\text{w},ij} = W_{\text{r},ji} \sim \mathcal{N}\left(0, \frac{\rho_2^2}{n}\right),
    \quad
     i \ne j
\end{equation}

\change{Wigner matrices are symmetric. In this work, we will always set $2 \rho^{}_1 = \rho^{}_2 = \rho$.
Notably, this leads to $ \langle \rho(\matr{W}_\text{r}) \rangle = \langle \sigma_{\text{max}}(\matr{W}_\text{r}) \rangle= \rho$.} 

%
\begin{figure*}
    \centering
    \includegraphics[width=\linewidth]{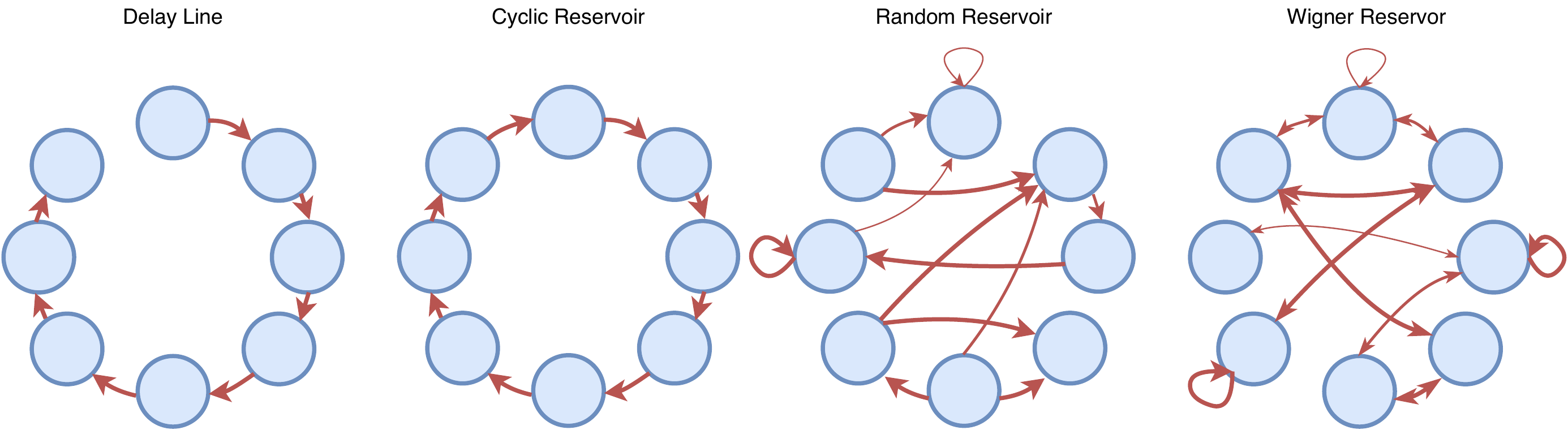}
        \caption{The different architectures discussed in this work. From left to right: delay line, cyclic, random and Wigner topology. The thickness of the arrow accounts for the strength of the connection. Notice that it has the same value for all the connections in both the delay line and the cyclic reservoir, while it varies for the other two. Note the presence of self loops in random and Wigner architectures. \change{Also consider that the Wigner reservoir has symmetric connections (double-headed arrows).}}
\label{fig:architectures}
\end{figure*}


\section{Controllability matrix and network encoded input} \label{sec:controllability}

\change{Here we develop a representation for the network state evolution based on the \ac{CH} theorem \cite{bernstein2009matrix}.}
The \ac{CH} theorem states that every real square matrix satisfies its own characteristic equation, implying that
\begin{equation}\label{eqn:an-linear-comb}
    \matr{W}^n = \varphi_{n-1} \matr{W}^{n-1} + \varphi_{n-2}\matr{W}^{n-2} + \dots + \varphi_{1} \matr{W} + \varphi_{0} \matr{I}    
\end{equation}{}
where the $\varphi_i$ are the \change{negated} coefficient of the characteristic polynomial  (see Appendix~\ref{app:CHT} for details).
Accordingly, any power of matrix $\matr{W}$ can be written as a linear combination of the first $n-1$ powers, where $n$ is the matrix order (and also the size of the reservoir):

\begin{equation}\label{eqn:CH_representation}
    \matr{W}^k = \sum_{j=0}^{n-1} \phi_j^{(k)} \matr{W}^j 
\end{equation}{}
where the apex $k$ denotes the fact that the $n$ coefficients are expansion coefficients  of the $k$-th power of $\matr{W}$.
In Appendix~\ref{app:input} we also show how the coefficients $\phi_j^{(k)}$ can be written in terms of $\varphi_{j}$ in \eqref{eqn:an-linear-comb}.

By \change{inserting} \eqref{eqn:CH_representation} in \eqref{eqn:res_general}, we obtain:
\begin{align} \label{eqn:state_sum}
    \vec{x}_0 &= \sum_{k=0}^\infty \sum_{j=0}^{n-1} \phi_j^{(k)} \matr{W}^j \vec{w}u_{-k} \\ &=\sum_{j=0}^{n-1}\matr{W}^j \vec{w} \sum_{k=0}^\infty \phi_j^{(k)}u_{-k} = \sum_{j=0}^{n-1}\matr{W}^j \vec{w} s_j
\end{align}{}
where 
\begin{equation} \label{eqn:s}
s_j := \sum_{k=0}^\infty \phi_j^{(k)}u_{-k}     
\end{equation}
is what we call the \emph{network encoded input}.
It is useful to interpret $\vec{s} = (s_0, s_1, \dots, s_{n-1})$ as a vector with $n$ components, which ``encodes'' the left-infinite input signal $\vec{u}$ in the spatial representation provided by the network.
In order for the $s_j$ term to exist, the sum \change{in \eqref{eqn:s}} must converge; we will discuss this issue in the next section.
We emphasize the fact that the sum over $j$ (the dimensionality of our system) is a \emph{finite} sum with $n$ terms, as opposed to the infinite sum over the $k$ (the time index).

Inspired by well-known tools from control theory \cite{sontag2013mathematical}, we define the \emph{controllability matrix} of the reservoir as
\begin{equation}
    \matr{\mathcal{C}} = [\vec{w} \quad \matr{W}\vec{w} \quad \matr{W}^2\vec{w} \quad \dots \quad\matr{W}^{n-1}\vec{w}  ]
\end{equation}{}

Then, the state-update equation \eqref{eqn:res_general} becomes
\begin{equation} \label{eqn:final_state}
    \vec{x}_{0} =  \matr{\mathcal{C}}  \vec{s}
\end{equation}{}
and the output \eqref{eqn:readout} can then be expressed as:
\begin{equation} \label{eqn:final_readout}
    y_0 = \vec{r} \matr{\mathcal{C}}  \vec{s}
\end{equation}{}
i.e., the readout filters the input according to controllability matrix.


\section{How the network encodes the input signal} \label{sec:encoded_input}

From \eqref{eqn:final_readout}, we see that the possibility for the readout to produce the \change{correct output (i.e., the output that solves the task at hand} depends on two distinct elements: the controllability matrix $\matr{\mathcal{C}}$ (function of $\matr{W}$ and $\vec{w}$) and the network encoded input $\vec{s}$ (which depends on $\matr{W}$ and  $\vec{u}$).

In Appendix~\ref{app:input} we show that the coefficients $\phi^{(k+1)}_{i}$ of \eqref{eqn:CH_representation} can be recursively expressed in terms of $\phi^{(k)}_{i}$ as:
\begin{equation}
    \begin{bmatrix}
           \phi^{(k+1)}_{0}\\
            \phi^{(k+1)}_{1}\\
         \vdots \\
           \phi^{(k+1)}_{n-2}\\
           \phi^{(k+1)}_{n-1}\\    
    \end{bmatrix}
    =
    \matr{M}
    \begin{bmatrix}
       \phi^{(k)}_{0}\\
       \phi^{(k)}_{1}\\
    \vdots \\
       \phi^{(k)}_{n-2}\\
       \phi^{(\change{k})}_{n-1}\\
    \end{bmatrix}
\end{equation}
where $\matr{M}$ is the Frobenius companion matrix of $\matr{W}$ (see Apprendix~\ref{app:input} for details). 
Note that the characteristic polynomial of $\matr{M}$ is that of $\matr{W}$; as such, the two matrices share the same eigenvalues.
Thus, the series \eqref{eqn:s} converges, for bounded inputs, when \change{$\matr{W}$} has a spectral radius smaller than $1$.\footnote{
\change{Note that most theoretical results (see e.g.,\cite{jaeger2001echo,grigoryeva2018echo, tino2020dynamical}) require the \ac{MSV} to be smaller than one, which is a stricter condition (as \ac{SR}$\le$\ac{MSV}); it follows that our analysis can be applied to a larger class of reservoirs.
}
}

The $\vec{s}$ vector can be written as:
\begin{equation}\label{eqn:s_general}
    \begin{bmatrix}{}
    s_0 \\
    s_1 \\
    \vdots \\
    s_{n-2}\\
    s_{n-1}
    \end{bmatrix}
    =
    \begin{bmatrix}
    \sum_{k=0}^\infty \phi_0^{(k)}u_{-k}   \\
    \sum_{k=0}^\infty \phi_1^{(k)}u_{-k} \\
     \vdots \\
    \sum_{k=0}^\infty \phi_{n-2}^{(k)}u_{-k} \\
    \sum_{k=0}^\infty \phi_{n-1}^{(k)}u_{-k}\\
    \end{bmatrix}
\end{equation}{}

In Appendix~\ref{app:CHT} we show that \change{for $k<n$, $\phi^{(k)}_j = \delta_{kj}$ holds}.
We also note that terms corresponding to time-step $k=n$ follow from \eqref{eqn:an-linear-comb}. 
This means that \eqref{eqn:res_general} can be written as:

%
\change{
\begin{equation}\label{eqn:s_expanded}
\begin{split}
     \begin{bmatrix}{}
    &s_0 \\
    &s_1 \\
    & \dots \\
    &s_{n-2}\\
    &s_{n-1}
    \end{bmatrix}
        &=
    \begin{bmatrix}
    &u_{0}  \\
    &u_{-1} \\
    &\dots \\
    &u_{-(n-2)}\\
    &u_{-(n-1)}  \\
    \end{bmatrix}
    +
    \begin{bmatrix}
   &u_{-n} \varphi_0      \\
    &u_{-n} \varphi_1 \\
    &\dots \\
    &u_{-n} \varphi_{n-2} \\
    &u_{-n} \varphi_{n-1} \\
    \end{bmatrix}
    \\
    &+
    \begin{bmatrix}
    & \sum_{k=n+1}^\infty \phi_0^{(k)}u_{-k}   \\
    &\sum_{k=n+1}^\infty \phi_1^{(k)}u_{-k} \\
    & \dots \\
    & \sum_{k=n+1}^\infty \phi_{n-2}^{(k)}u_{-k} \\
    & \sum_{k=n+1}^\infty \phi_{n-1}^{(k)}u_{-k} \\
    \end{bmatrix}
\end{split}
\end{equation}
}

All other terms in \eqref{eqn:s_expanded} corresponding to time steps $k>n$ can be computed according to \eqref{eqn:CH_representation}.

This procedure shows that, in general, the inputs from $0$ to $n-1$  steps back in time will \emph{always} appear in their original form, and the cross-contribution  starts only  from  $u^{}_{-n}$ backwards in time. 
We will make use of this result to analytically examine the properties of different networks topologies. 
Moreover, by deriving the expression for the $\phi_{i}^{(k)}$ we can study how the network is able to recall its past inputs. 
In general, if the $\phi_{i}^{(k)}$s are large then the network will not be able to recall the inputs, since the input $u_{-j}$  can only be read through $s_j = u_{-j} + \sum_{k=n}^\infty \phi_j^{(k)}u_{-k}$. 
It follows that having large expansion coefficients  $\phi_j^{(k)}$ prevents the network from being able to recall its past inputs.
We will show in the next section that, when we can derive an analytical expression for the $\phi_{i}^{(k)}$, it is possible to anticipate how the network recalls its past inputs.
Note that, as implied by \eqref{eqn:final_readout}, for a linear network this is deeply related to its expressive power, since the network output is basically a linear combination of past inputs.
The inputs accessibility  to the readout is also due to $\matr{\mathcal{C}}$, which is a property of the network only, as it does not depend on any input signal.
\change{A detailed discussion about the relation between the network properties and the rank of the controllability matrix $\matr{\mathcal{C}}$ was recently presented in \cite{gonon2020memory}. There, the authors prove that the memory capacity for linear reservoirs equals the rank of $\matr{\mathcal{C}}$. 
In the following, we analyze the different architectures described in Section~\ref{sec:RC} discussing the properties of their controllability matrices.
}

In the random \change{reservoir}  case (Subsection~\ref{subsec:random}, the property of $\matr{\mathcal{C}}$ can be studied by considering the expected values of the norm of its columns,\change{which describes how the system accesses past inputs.}
\change{Let us consider the $n$-by-$n$ matrix $\matr{W}_r$ in (9) and a vector with $n$ components $\vec{w} = \{  w_{j} \} \sim \mathcal{N}(0,\frac{1}{n})$.
     {Since $w_j$ are generated independently, the expected value of the squared norm of the random vector $\vec{w}$} is $l(\vec{w}) = n \langle w_i^2 \rangle$.
    We drop the $r$ in $\matr{W}$, to simplify the notation.
    We now study  $\vec{z} := \matr{W}\vec{w}$ and obtain:
    \begin{equation}
        \langle z_i^2 \rangle 
        = \langle (\matr{W}\vec{w})_i^2 \rangle 
        = \langle (\sum_j W_{ij}w^{}_j)^2   \rangle
        = n\langle W_{ij}^2 \rangle \langle w_i^2 \rangle
    \end{equation}{}
    where the last equality follows from the independence of the  {zero-mean} entries of  $\matr{W}$ and $\vec{w}$.
    Now, by the way we constructed $\matr{W}$ and $\vec{w}$, we see that $\langle W_{ij}^2 \rangle = \rho^2/n$ and that $\langle w_i^2 \rangle = 1/n$. This results in:
    \begin{equation}
        \langle z_i^2 \rangle =  n \langle W_{ij}^2 \rangle \langle w^2 \rangle =  n \frac{\rho^2}{n}  \frac{1}{n} = \frac{\rho^2}{n}
    \end{equation}{}
    This means that $l(\vec{z}) = n \langle z_i^2 \rangle = \rho^2$ and that the  standard deviation is $\sqrt{\langle z_i^2 } \rangle = \frac{\rho}{\sqrt{n}}$.\\}
From the above the first column of $\matr{\mathcal{C}}$ has  euclidean norm $\lVert \vec{w} \rVert^{} = 1$, the second $\rho$ , the third $\rho^2$; the last one $\rho^{(n-1)}$. 
Since $\rho$ must be smaller than $1$, the \change{components of} last columns of $\matr{\mathcal{C}}$ \change{shrink quickly}.
This fact explains the \change{shading} observed in the column of $\matr{\mathcal{C}}$ for the random case of Fig.~\ref{fig:C_matrix} and Fig.~\ref{fig:C_matrix1000}.
For the Wigner case the effect if emphasized by the correlations introduced by the symmetry of $\matr{W}_\text{w}$.
The controllability matrix $\matr{\mathcal{C}}$ for the delay line and the cyclic reservoir can instead be described in exact terms  (see Appendices~\ref{app:delay_line} and \ref{app:cyclic}).
A sample of each case in provided in Fig.~\ref{fig:C_matrix} and Fig.~\ref{fig:C_matrix1000} for $n=100$ and $n=1000$, respectively.
For the delay line a  complete analysis of the network output can be carried.
As shown in  Appendix~\ref{app:delay_line}, we can write that:
\\
\begin{equation}
    \vec{y}_0 
    = \vec{r} \cdot \matr{I} \cdot \vec{s}_\text{d} = \sum_{i=0}^{n-1} r_i u_{-i}
\end{equation}{}
\\
where $\matr{I}$ is the \emph{identity matrix}.
This is, as one would expect, simply a regression model of order $n$.

For the cyclic reservoir architecture (Subsection~\ref{subsec:cycli}) we  show in  Appendix~\ref{app:cyclic} that defined the $i$-time permuted input weights vector as:
\begin{equation}
    \vec{w^{(i)}} := \matr{W}_\text{c}^i \vec{w}
\end{equation}{}
then, the output of the cyclic reservoir at time zero $y_0$ is:
\begin{equation}\label{eqn:output_final}
    y_0 = \vec{r} \cdot \Tilde{\matr{\mathcal{C}}}_\text{c} \cdot \Tilde{\vec{s}}
\end{equation}
where
\begin{align}
\label{eqn:s_hat}
    &\Tilde{s}_j = \sum_{p=0}^{\infty} \rho^{j+pn} u_{-{j+pn}} \\
    \label{eqn:C_hat}
    &\Tilde{\matr{\mathcal{C}}}_\text{c} = [\vec{w} \quad \vec{w}^{(1)} \quad \vec{w}^{(2)} \quad \dots \vec{w}^{(n-1)} ]
\end{align}{}

The fact that, as suggested in \cite{rodan2010minimum}, $\vec{w}$ should be non-periodic for the network to work at its best, is now evident. 
In fact, if $\vec{w}$ is periodic, it means that some columns of $\hat{\matr{\mathcal{C}}}_\text{c}$ are linearly related and, therefore, the rank degenerates, as supported by theoretical arguments in \cite{tino2020dynamical}.

Note that $u_{-j}$ is only readable through the term \change{$s_{j} = u_{-j} + \rho^n u_{-(j+n)} + \dots$} and, in order to do that, it must \change{hold} that \change{$u_{-j} \gg \rho^n u_{-(j+n)}$}. 
This may suggest to choose small spectral radii, but the smaller the spectral radius, the faster the decay of the memory, since \change{$\hat{s}_j = \rho^{j} s_j$}. 
This confirms previous intuitions \cite{rodan2010minimum}, stating that by choosing a small spectral radius, the network preserves an accurate representation of recent inputs, at the expense of losing the ability to recall remote ones. 
Conversely, if one sets a large spectral radius (i.e., close to $1$) the network will be able to (partially) recall inputs from the past, but its memory of more recent inputs will decrease.


\section{The nullspace of C and the network memory}\label{sec:null_space}

By \eqref{eqn:final_readout} one can understand how the rank of the controllability matrix $\matr{\mathcal{C}}$ is associated with the degrees of freedom (the effective number of parameters used by the model to solve the task at hand) that can be exploited by the readout  (i.e., the ``complexity'' of the model). 

\begin{figure*}
    \centering
    \includegraphics[width = \textwidth]{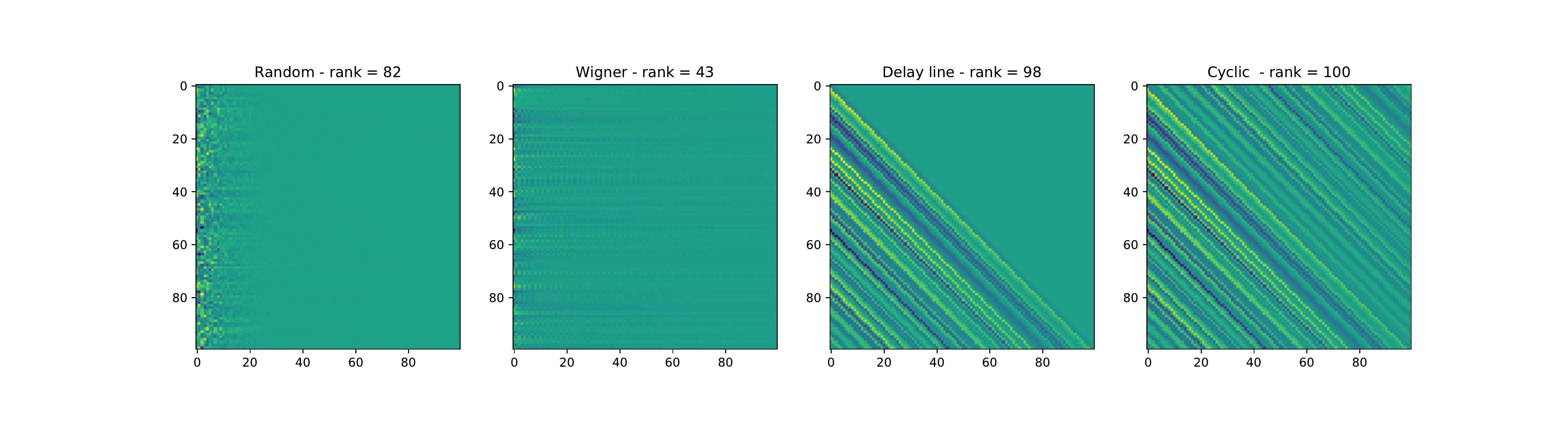}
    \caption{The controllability matrix and its rank for different architectures. The Spectral Radius is $\rho = 0.99$ and the reservoirs has $N = 100$ neurons. The four architectures share the same randomly-generated $\vec{w}$.}
    \label{fig:C_matrix}
\end{figure*}

\begin{figure*}
    \centering
    \includegraphics[width =\textwidth]{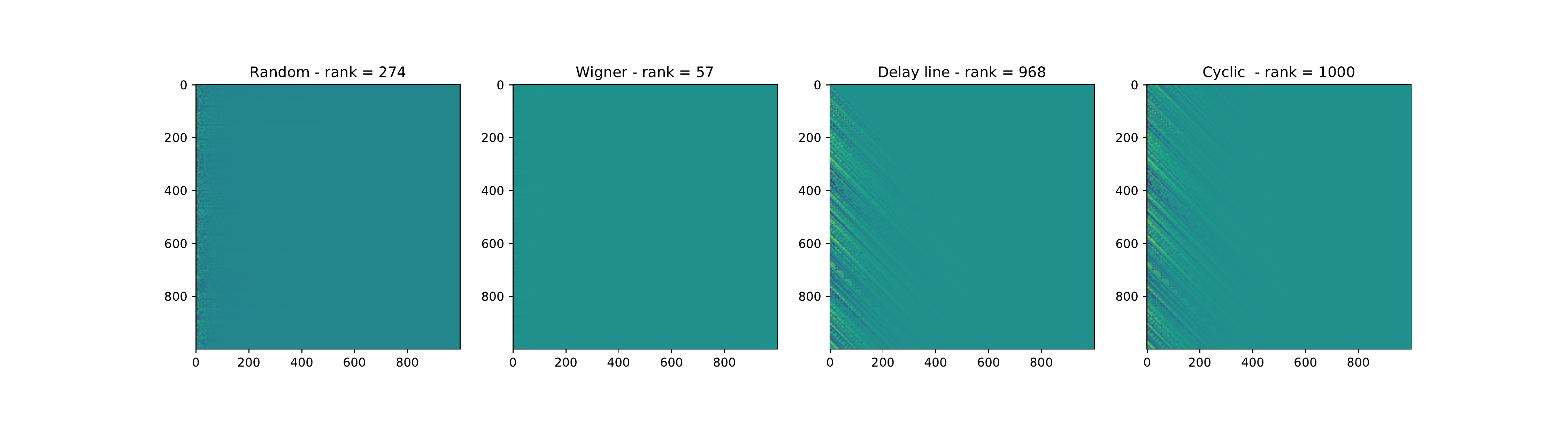}
    \caption{The controllability matrix and its rank for different architectures. The Spectral Radius is $\rho = 0.99$ and the reservoirs has $N = 1000$ neurons. The four architectures share the same randomly-generated $\vec{w}$.}
    \label{fig:C_matrix1000}
\end{figure*}

Note from Figures~\ref{fig:C_matrix} and \ref{fig:C_matrix1000} that the cyclic reservoir always has the highest rank of $\matr{\mathcal{C}}$, while the Wigner the lowest. The difference increases with the number of neurons.\footnote{Note that this is coherent with the findings in \cite{tino2020dynamical}, since the $\matr{Q}$ defined in that work is simply $Q =\matr{\mathcal{C}}^{\top} \matr{\mathcal{C}}$ and the number of motifs is related to the rank of $\matr{Q}$ (and so, of $\matr{\mathcal{C}}$).}
{}
{}

\begin{figure}[h]
    \centering
        \begin{subfigure}{\linewidth}
            \centering
        \includegraphics[width = .75\linewidth]{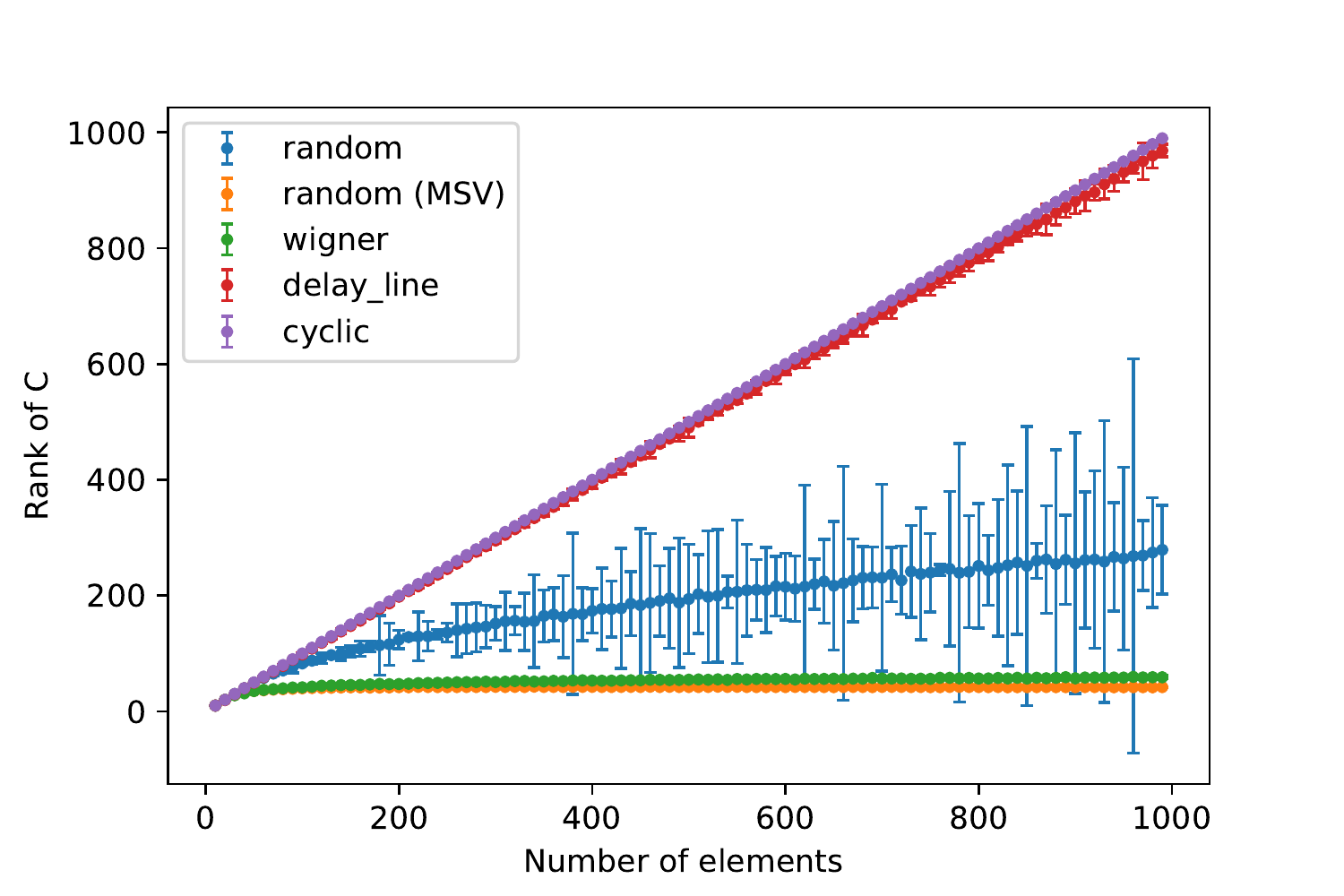}
        \caption{}
        \label{subfig:ranks}
        \end{subfigure}
        
        \begin{subfigure}{\linewidth}
            \centering
        \includegraphics[width = .75\linewidth]{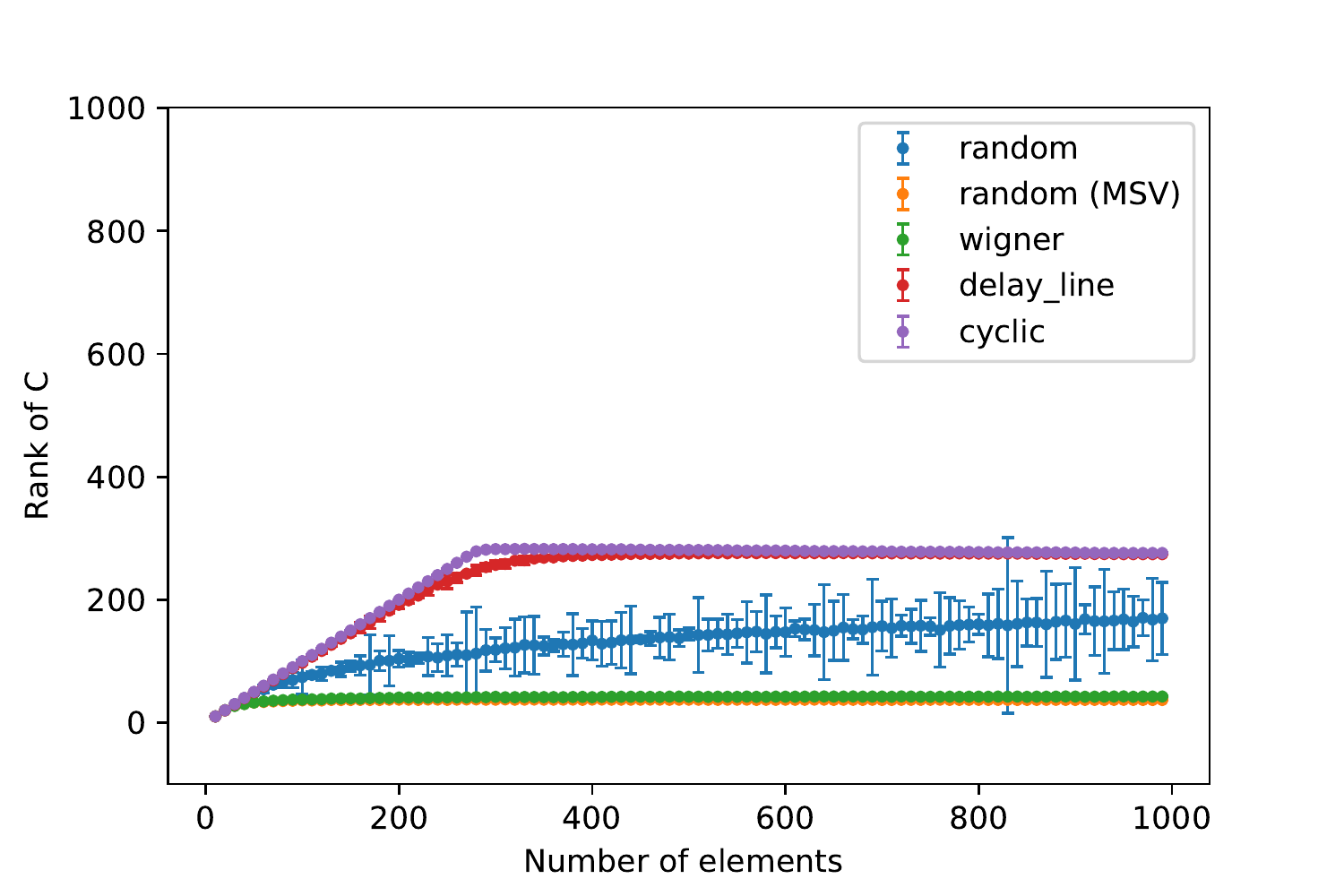}
        \caption{}
        \label{subfig:ranks_smallerSR}
        \end{subfigure}
 \caption{Ranks of the controllability matrix  $\matr{\mathcal{C}}$ as a function of the reservoir dimension $n$, for $\rho=0.995$ (\protect\subref{subfig:ranks}) and $\rho=0.9$ (\protect\subref{subfig:ranks_smallerSR}). 
 \change{Note the saturation of the delay line and the cyclic reservoir, which happens, because $\rho$ is not close enough to $1$ for its powers to be numerically distinguishable from zero.
 }
 }
    \label{fig:ranks_total}
\end{figure}

\begin{figure}[h]
    \centering
        \begin{subfigure}{\linewidth}
            \centering
        \includegraphics[width =.5 \linewidth]{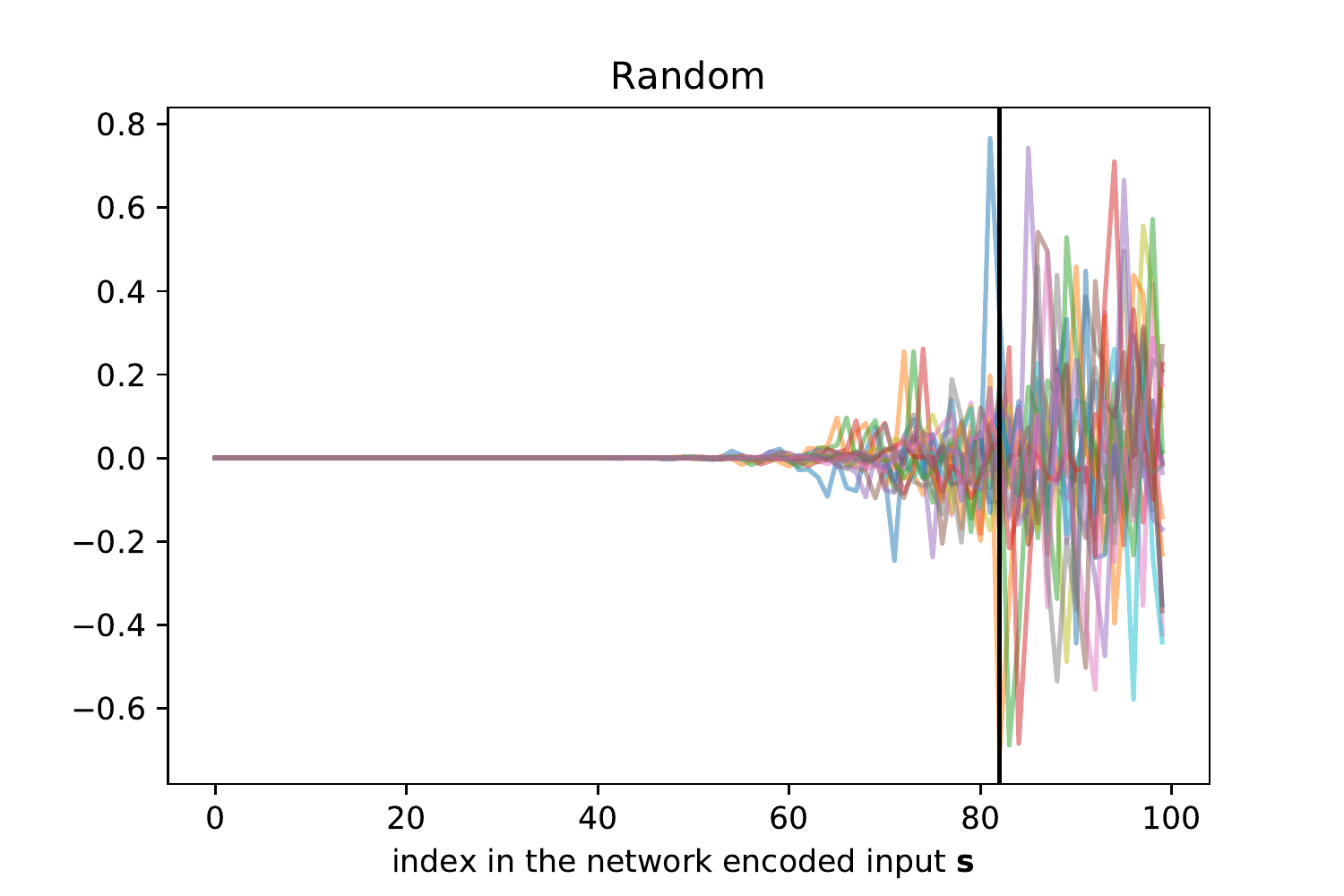}
        \caption{}
        \label{subfig:random_nullspace}
        \end{subfigure}
        \\
        \begin{subfigure}{\linewidth}
            \centering
        \includegraphics[width =.5 \linewidth]{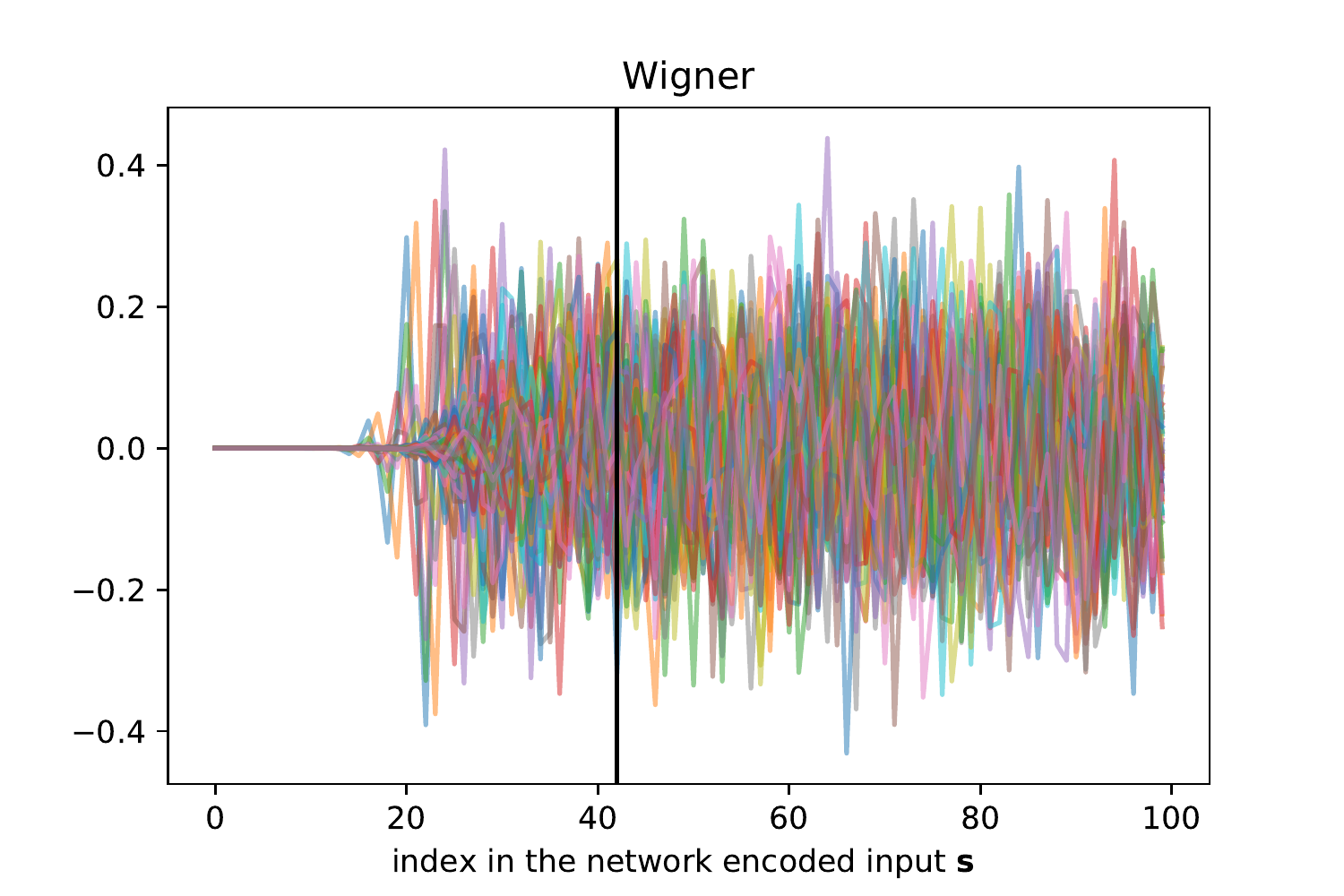}
        \caption{}
        \label{subfig:wigner_nullspace}
        \end{subfigure}
    \caption{Nullspace basis for a Random (\protect\subref{subfig:random_nullspace}) and a Wigner (\protect\subref{subfig:wigner_nullspace}) reservoir matrices.
    \change{Each curve represents a basis vector of the nullspace of $\matr{\mathcal{C}}$, where the $x$-axis accounts for the vector components.
     In both cases, we set the spectral radius to $\rho = 0.99$, while the reservoirs size is $n = 100$.
     The vertical black lines represent the ranks of $\matr{\mathcal{C}}$, i.e., the dimensions of the image spaces. 
     }
     }
    \label{fig:nullspace}
\end{figure}

The fact that $\matr{\mathcal{C}}$ is not full-rank is linked to the presence of the \emph{nullspace}.%
\footnote{What we call the nullspace is practically the \emph{effective} nullspace detected up to the numerical precision, computed using the Numpy dedicated function \cite{oliphant2006guide}.}
This means that there are some network encoded inputs $\vec{s}$ which are mapped to $\vec{0}$ by $\matr{\mathcal{C}}$ \eqref{eqn:final_state} and hence are indistinguishable by the readout perspective. 
In Fig.~\ref{fig:ranks_total}, we plot the rank of $\matr{\mathcal{C}}$ as a function of the reservoir dimension $n$.
In the experiments using Wigner and cyclic reservoirs, the spectral radius $\rho$ and the maximum singular value $\sigma_{\text{max}}$ coincide and their values are set to $0.995$ (Fig.~\ref{subfig:ranks}) and $0.9$ (Fig.~\ref{subfig:ranks_smallerSR}).
For the random reservoir, $\rho$ and $\sigma_{\text{max}}$ are distinct, so we design an experiment where the spectral radius is fixed and another one where the maximum singular value is set (we remind the reader that $\langle \rho \rangle = \frac{1}{2}\langle \sigma_{\text{max}} \rangle$). 
But what is the shape of the basis of this nullspace? 
We show its basis in two cases (see Fig. \ref{fig:nullspace}).
The controllability matrix obtained with a cyclic reservoir does not have a nullspace for such a value of $\rho$, since $\matr{\mathcal{C}}$ is full-rank.
Note that, in order to interpret each vector in Fig. \ref{fig:nullspace} as a time series, one must consider the last inputs seen as the ones closer to the origin. 
Given this interpretation, we clearly see how the memory is linked to the rank of $\matr{\mathcal{C}}$: the reservoirs' ability to recall past inputs depend on the rank of $\matr{\mathcal{C}}$ as inputs which only differ in the far-away past are mapped to the same final state.
{}

Let us consider an example.
Let $\vec{s}_1$ and $\vec{s}_2$ be two network encoded inputs, which differ only in the last $n-m$ elements.
Denote by $\vec{x}^i_0$ the final state of the system after being fed by the signal $\vec{s}_i$. 
We can then write $\vec{s}_2 =  \vec{s}_1 + \vec{d}$, where $\vec{d}$ encodes the difference between the two representations.
We see that the first $m$ elements of $\vec{d}$ are null.
So, we can write:
\begin{equation}
    \vec{x}^2_0 = \matr{\mathcal{C}} \vec{s}_2 = \matr{\mathcal{C}}(\vec{s}_1 + \vec{d})
    =  \matr{\mathcal{C}} \vec{s}_1 + \matr{\mathcal{C}} \vec{d} 
    = \matr{\mathcal{C}} \vec{s}_1 + \vec{0} = \vec{x}^1_0 
\end{equation}
since $\vec{d}$ lives in the nullspace of $\matr{\mathcal{C}}$. 
This results in the network not being able to distinguish between the two signals.


\section{Memory curves}\label{sec:memory_curves}
In order to validate our theoretical claims, \change{we designed networks asked to remember} a random i.i.d. input. 
We generate inputs of length $T$ and split them in a \emph{training set} ranging in $(0,t_0)$ and a \emph{test set} ranging in $(t_0,T)$. 
In the task under consideration, the network is trained to reproduce past input (a white noise signal) at a given past-horizon $\tau$, so that $y_k \equiv u_{k-\tau}$.
The input signal $\vec{u}$ is chosen to be Gaussian i.i.d. white noise, $u_{k} \sim \mathcal{N}(0,1)$.
With choice $T = 1500$ and $t_0 = 1000$,  the training set contains $L_{\text{train}}=1000$ sample and the test set $L_{\text{test}}=500$.
The readout is finally configured through a least means square procedure and its accuracy is then evaluated on the test set as $\gamma = \max\{1 - \text{NRMSE}, 0\}$, where the \ac{NRMSE}:
\begin{equation}\label{eqn:error}
    \text{NRMSE} := \sqrt
    {
    \frac{ \sum_{k=t_0}^T({y}_k - \hat{y}_k)^2}
    { \sum_{k=t_0}^T ( {y}_k - \overline{y})^2} 
    }
\end{equation}
${y}_k$ denotes the \change{system output} at time $k$, $\overline{y} := \frac{1}{T}\sum_{k=t_0}^T y_k$ is its average and $\hat{y}_k$ stands for the predicted output.

\begin{figure*}[h]
     \centering
     \begin{subfigure}[b]{0.45\textwidth}
         \centering
         \includegraphics[width=\textwidth]{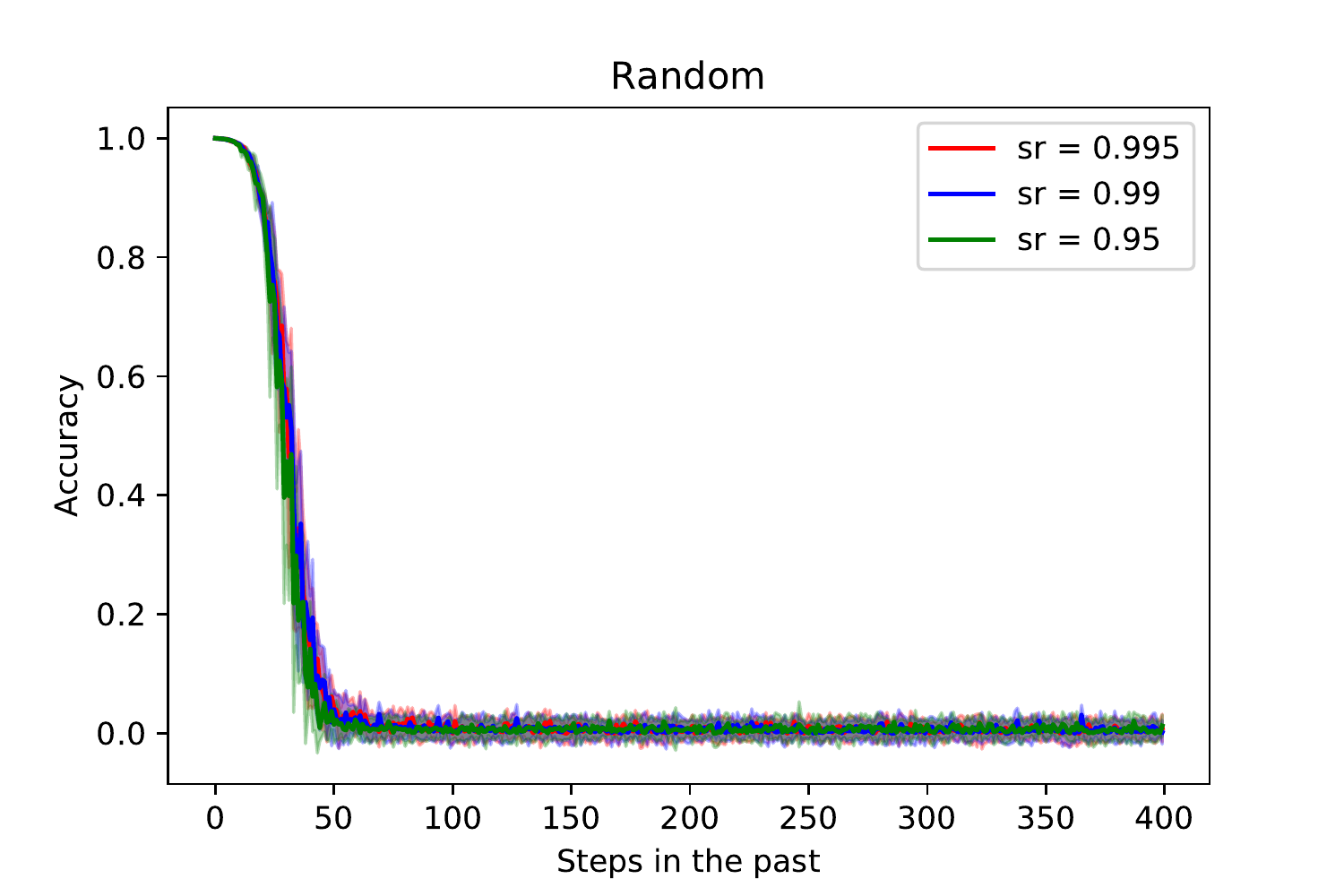}
         \caption{}
         \label{subfig:memory_curves_random}
     \end{subfigure}
     \hfill
     \begin{subfigure}[b]{0.45\textwidth}
         \centering
         \includegraphics[width=\textwidth]{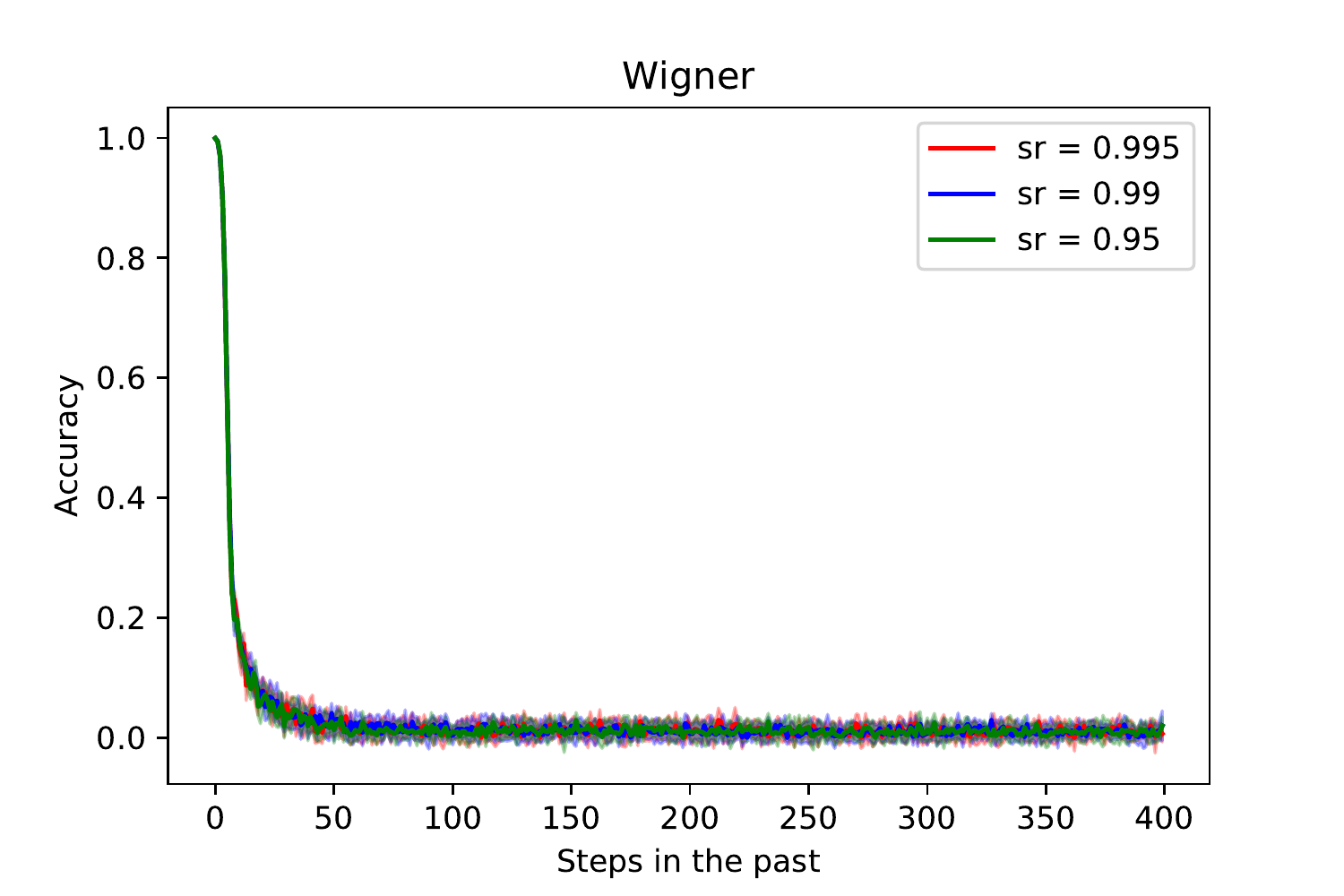}
         \caption{}
         \label{subfig:memory_curves_wigner}
     \end{subfigure}
    
    ~
    
    \begin{subfigure}[b]{0.45\textwidth}
         \centering
         \includegraphics[width=\textwidth]{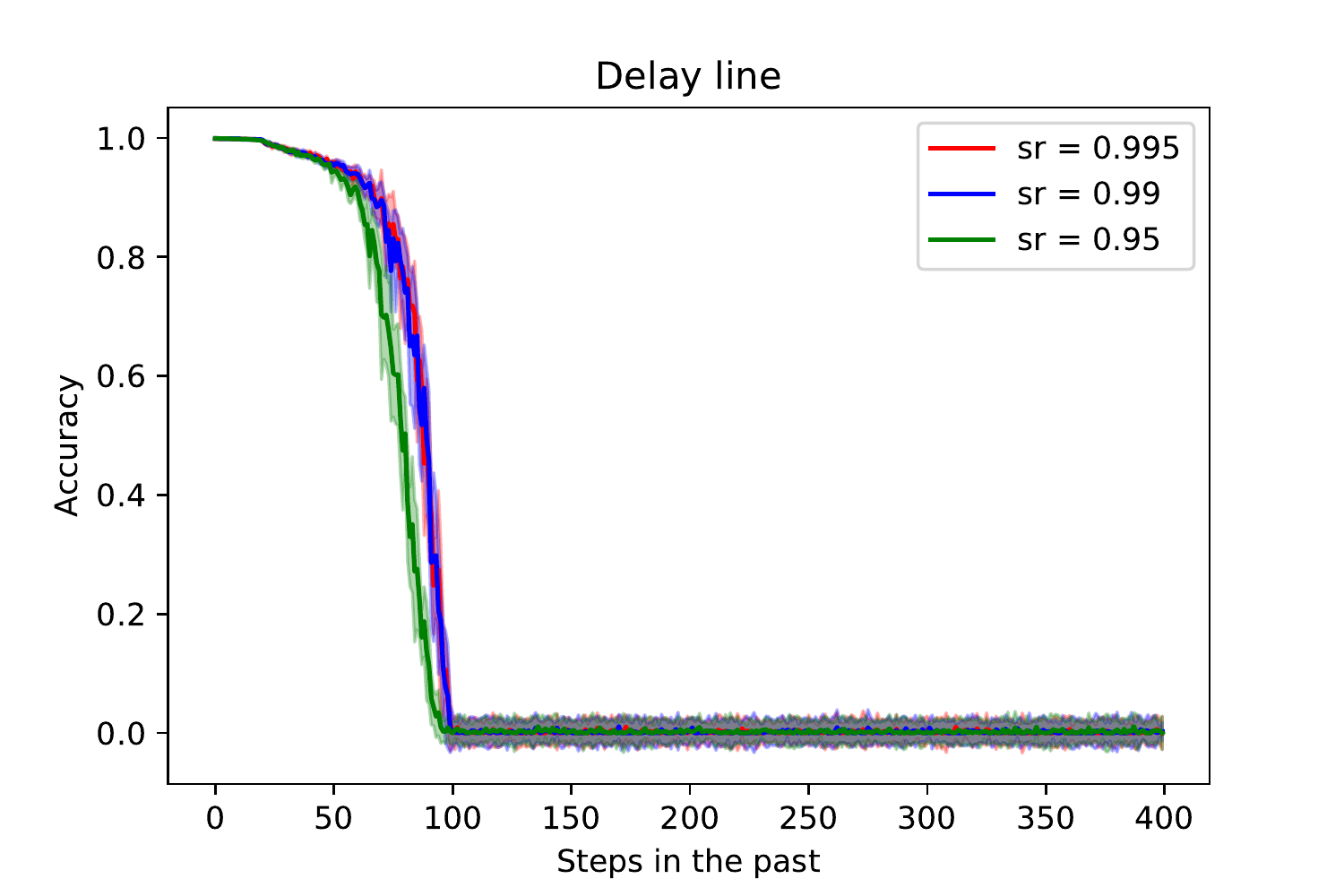}
         \caption{}
         \label{subfig:memory_curves_delay_line}
     \end{subfigure}
     \hfill
     \begin{subfigure}[b]{0.45\textwidth}
         \centering
         \includegraphics[width=\textwidth]{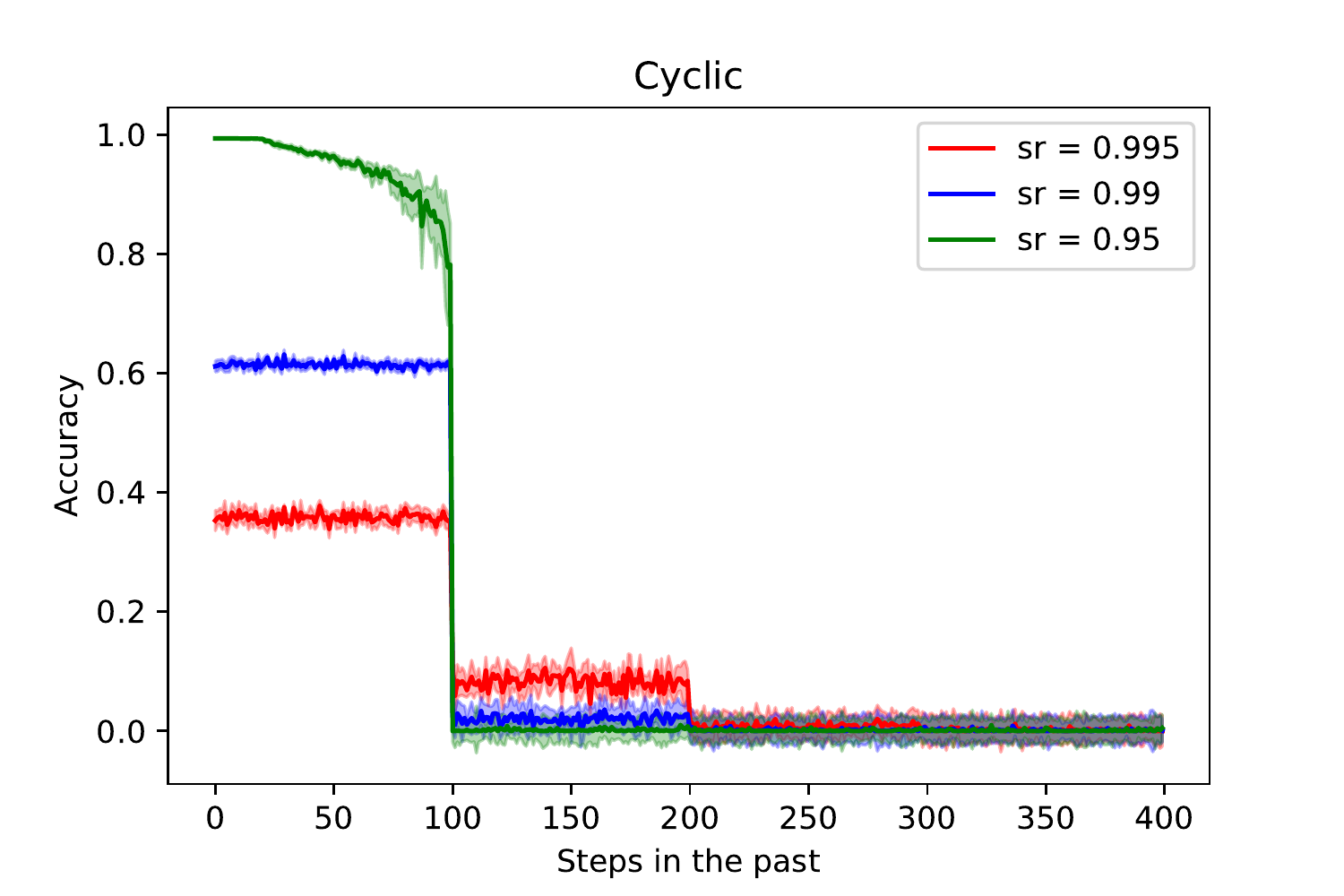}
         \caption{}
         \label{subfig:memory_curves_cyclic}
     \end{subfigure}
        \caption{Memory curves for the random, Wigner, \change{delay line} and cyclic reservoir with different values of $\rho$, for a reservoir of $N = 100$ neurons. 
        Plotted values are averages over $10$ different realizations, with the shaded area accounting for a standard deviation:
        \change{a side effect of plotting data in this way is that the value may be negative even if the accuracy is defined as a positive quantity.}
        Note that having a high ability to reconstruct recent inputs ($k < N$) compromises the capacity to remember the more distant ones. 
        }
        \label{fig:memory_curves}
\end{figure*}

In Figure~\ref{fig:memory_curves} the \emph{memory curves}, introduced in \cite{jaeger2001echo}, are plotted for four different architectures. 
The Random and the Wigner architectures appear to have a short memory. 
Their accuracy dramatically decreases as $\tau$ grows.
The behavior of the cyclic reservoir appears to be radically different. As described in \cite{rodan2010minimum}, the performance does not decrease gradually, but remains almost constant for some time and then abruptly decreases. 
The drop in performance occurs when $\tau = n$, \change{$n$ being the number of neurons in the reservoir.} 
This is coherent with the theory we developed and with the findings in \cite{rodan2012simple}.
\change{Note that in \cite[Section 3]{jaeger2001short} a similar shape for the memory curve is obtained by using an ``almost unitary'' reservoir matrix, where all singular values equal a constant $C<1$.
We note that also the cyclic reservoir $\matr{W}_c$ shares this feature, explaining why the results are similar.
}

We comment on how the \ac{SR} affects the performance: when the \ac{SR} is close to one, the accuracy \change{in the reconstruction of $\vec{u}(t - \tau)$} is lower for recent inputs samples (i.e., smaller $\tau$) but higher for distant in time ones.
In other words, choosing a large $\rho$ allows the network to better remember the distant past, at the price of compromising its ability to remember the recent one.
This  is a direct implication of \eqref{eqn:s_hat} and \eqref{eqn:s_expanded}: a larger \ac{SR} amplifies the contribution of past inputs over the more recent ones, since the \change{input reproducibility} property is controlled by $\rho^{j+pn}$ (see Appendix~\ref{app:input} for more details).

\begin{figure*}[h]
        \centering
        \begin{subfigure}[b]{0.45\textwidth}
         \centering
         \includegraphics[width=\textwidth]{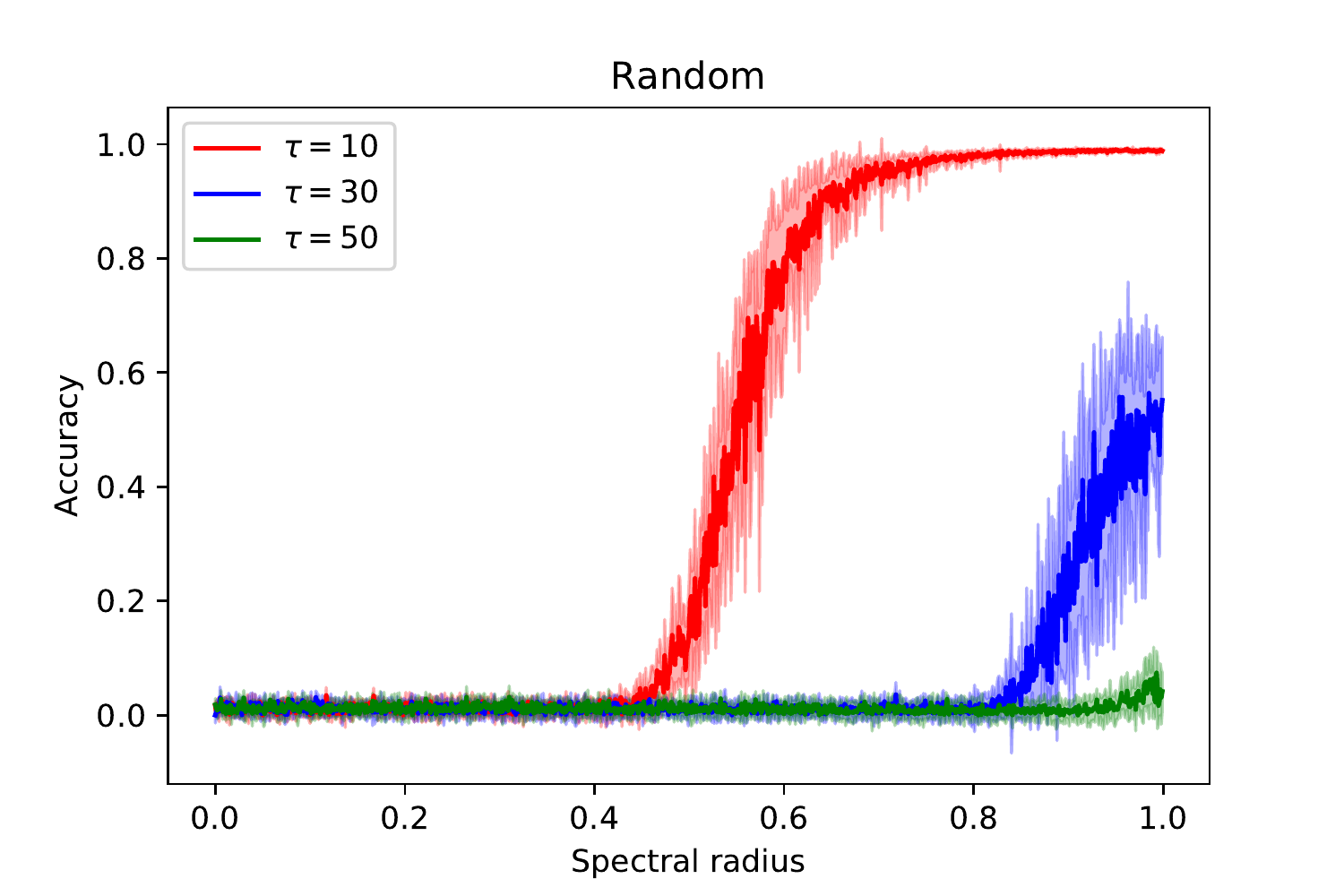}
         \caption{}
         \label{subfig:memory_SR_random}
        \end{subfigure}
        \hfill
        \begin{subfigure}[b]{0.45\textwidth}
         \centering
         \includegraphics[width=\textwidth]{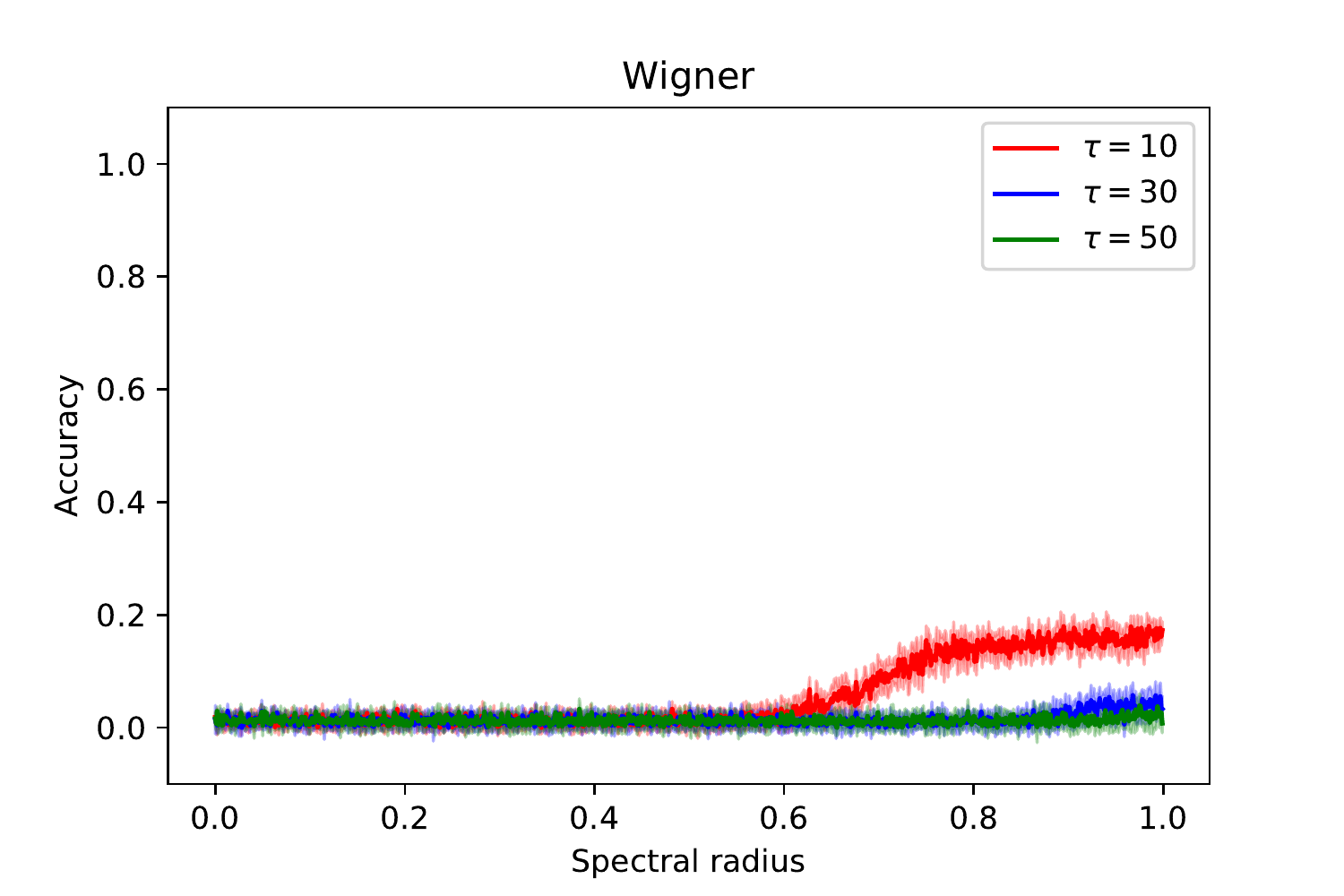}
         \caption{}
         \label{subfig:memory_SR_wigner}
        \end{subfigure}
        
        ~
        
        \begin{subfigure}[b]{0.45\textwidth}
         \centering
         \includegraphics[width=\textwidth]{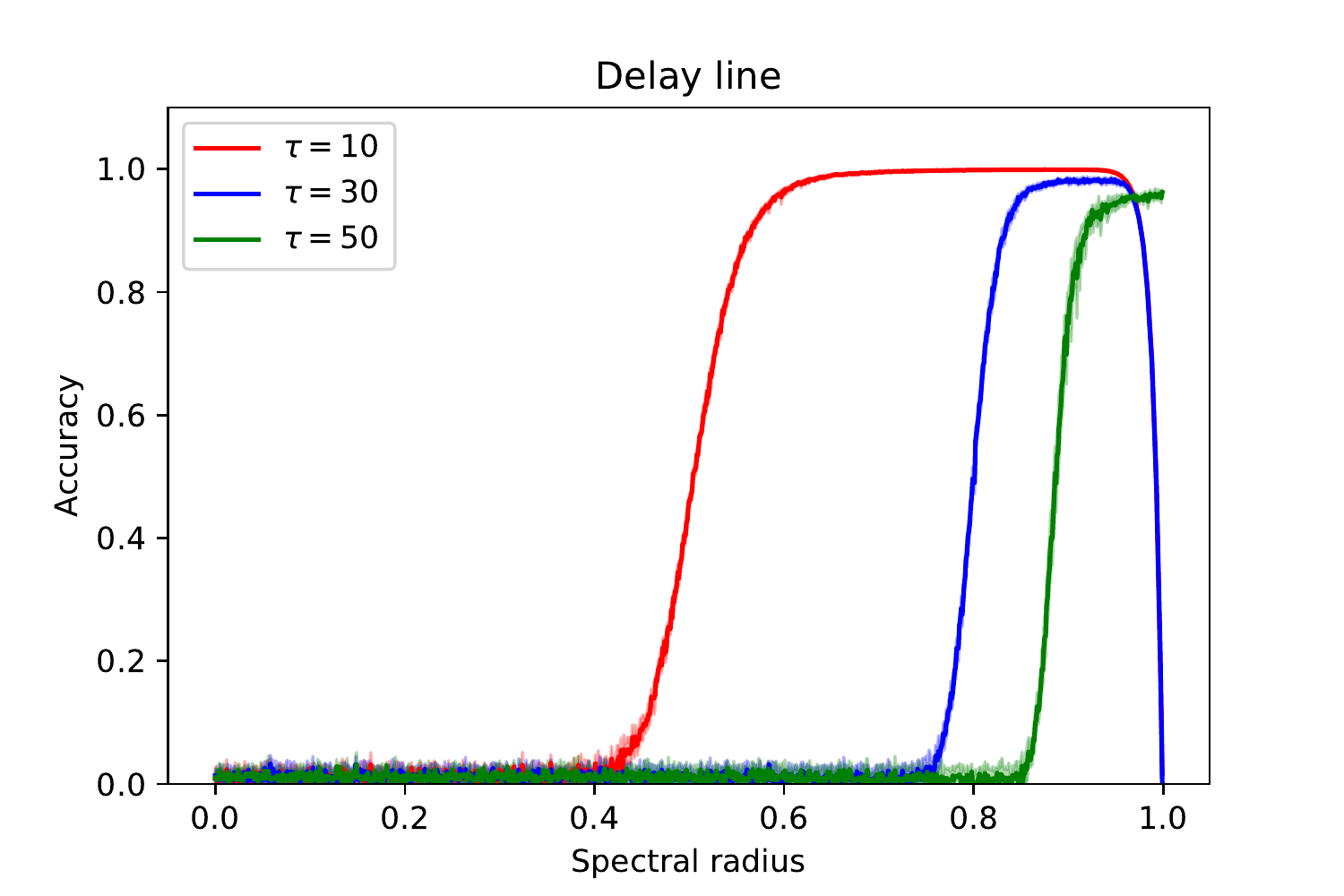}
         \caption{}
         \label{subfig:memory_SR_delay_line}
        \end{subfigure}
        \hfill
        \begin{subfigure}[b]{0.45\textwidth}
         \centering
         \includegraphics[width=\textwidth]{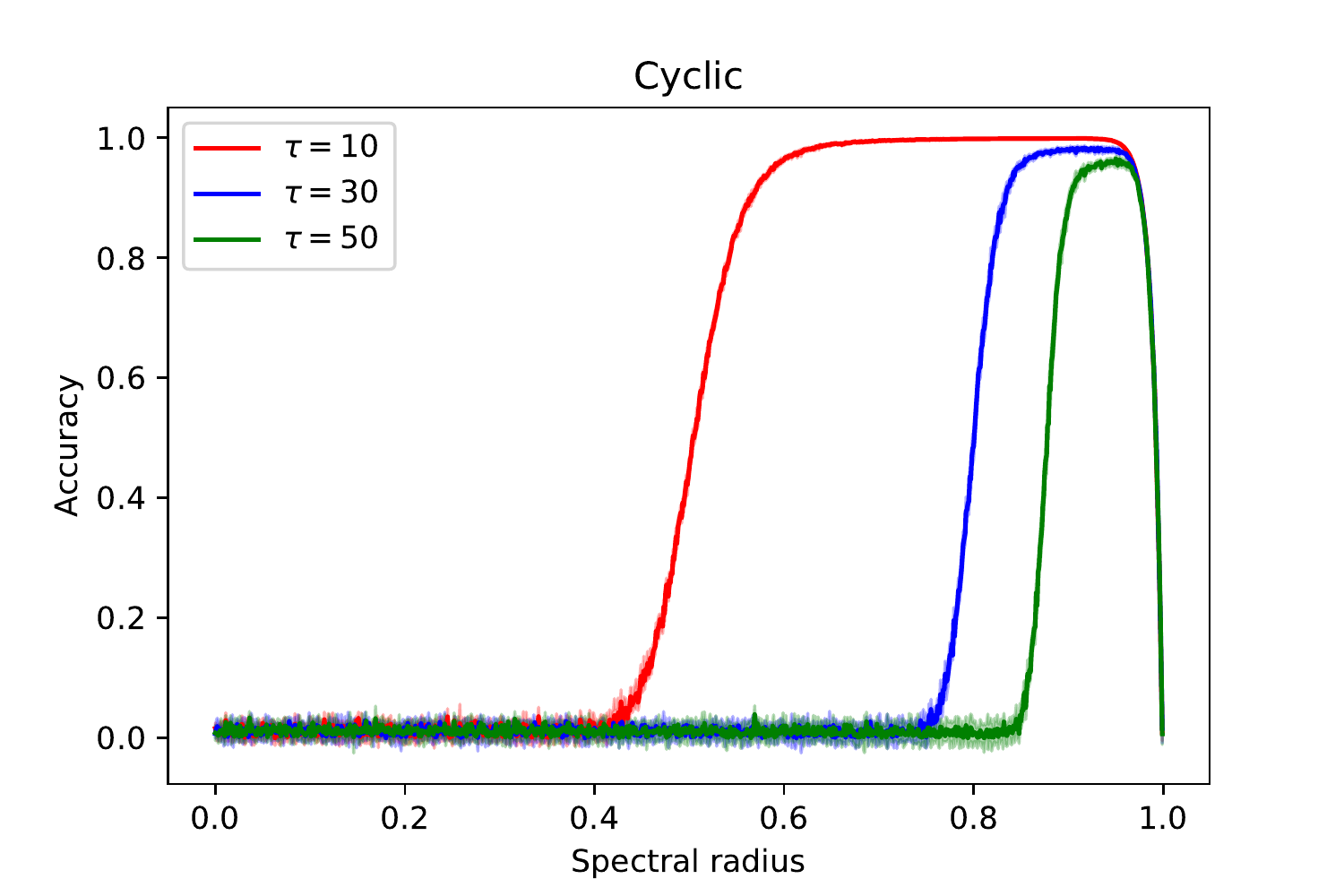}
         \caption{}
         \label{subfig:memory_SR_cyclic}
        \end{subfigure}
     
    \caption{Accuracy in remembering an i.i.d. past-input as function of the spectral radius. All the networks have $N = 100$ neurons.
     Plotted values are averages over $10$ different realizations, with the shaded area accounting for a standard deviation: \change{ a side effect of plotting data in this way is that the value may be negative even if the accuracy is defined as a positive quantity.}}
    \label{fig:memory_SR}
        
\end{figure*}

We investigate the impact of the spectral radius on memory capacity in Figure~\ref{fig:memory_SR}, where the accuracy $\gamma$ of the three architectures in recalling a past input (at various  $\tau$) is plotted as function of the \ac{SR}. 
According to our predictions, a larger \ac{SR} is required to correctly recall inputs that are further in past (but for which $\tau < N$), since the \ac{SR} controls the magnitude of the $\phi_j^{(k)}$, i.e., the permanence of $u_{k}$ on the state.
We notice that the Random and the Wigner architectures show a similar behavior, with the former displaying a superior performance than the latter. 
Instead, the Cyclic network has the same behavior as the $\ac{SR}$ increases, but displays an abrupt fall as it approaches $1$.
\change{
This is coherent with the theory we developed, since for $\rho \approx 1$ the powers of $\rho$  in \eqref{eqn:s_hat} do not shrink towards zero fast enough to forget remote inputs with the consequence that the network state will be an unreadable superposition of all the past outputs}

\section{Conclusions}\label{sec:conclusions}

In this paper, we proposed a methodology for explaining how linear reservoirs encode inputs \change{in their internal states}.
Theoretic findings allow to  express the system state in terms of the \emph{controllability matrix} $\matr{\mathcal{C}}$ and the \emph{network encoded input} $\vec{s}$.
The matrix properties of $\matr{\mathcal{C}}$ allow us to compare different connectivity patterns for the reservoir in a quantitative way by.
Results show that reservoirs with a cyclic topology give the richest possible encoding of input signals, yet they also offer one the most parsimonious reservoir parametrization.
To the best of our knowledge, our contribution pioneers the rigorous study on how specific coupling patterns for the recurrent layer and individual setting of the dynamical system influence computational properties (e.g., memory), providing deeper insights about phenomena that so far have been observed only empirically in the literature.


%

\appendix
\section{Cayley-Hamilton Theorem }\label{app:CHT}

The \ac{CH} Theorem allows one to describe the $n$-th power of a matrix in term of the first $n-1$-powers (including the zero power, which is the identity).

Let $\matr{W} \in \mathbb{R}^{n \times n}$ be a square matrix. 
Its characteristic polynomial is defined as:
\begin{equation}
    \det{ (\lambda \matr{I} - \matr{W})} = 0 \quad \rightarrow \quad \lambda^n  + \alpha_{n-1}\lambda^{n-1} + \dots + \alpha_{1} \lambda+ \alpha_{0} = 0  
\end{equation}
where $\lambda$ is an eigenvalue of $\matr{W}$ and the $\alpha_{k}$ are the coefficients of the characteristic polynomial.

\begin{theorem}[Cayley-Hamilton]\label{theor:CH}
Every real square matrix satisfies its characteristic equation
\begin{equation}
    \matr{W}^n + \alpha_{n-1}\matr{W}^{n-1} + \dots + \alpha_{1}\matr{W}^{1} + \alpha_{0}\matr{I} = \matr{0}. 
\end{equation}{}
\end{theorem}{}
Accordingly, it is possible to show that the $n$-th power of the matrix can be represented as a \emph{linear combination} of its lower powers:
\begin{equation}
\label{eq:matrix_power}
    \matr{W}^n  = -\alpha_{n-1}\matr{W}^{n-1} - \dots - \alpha_{1}\matr{W}^{1} - \alpha_{0}\matr{I}  
\end{equation}{}

For  matrix $\matr{W}$, Theorem \ref{theor:CH} states:
\begin{equation}
\label{eq:an-linear-comb}
    \matr{W}^n = \varphi_{n-1} \matr{W}^{n-1} + \varphi_{n-2}\matr{W}^{n-2} + \dots + \varphi_{1} \matr{W} + \varphi_{0} \matr{I}
\end{equation}{}
Here, $\matr{W}$ is a $n\times n$ matrix, $\matr{I}$ is the $n\times n$ identity matrix, and $\varphi_{k}=-\alpha_k$ are the negated coefficients of the characteristic polynomial of $\matr{W}$.
It holds true that
\begin{equation}
    \matr{W}^m = \phi_{n-1}^{(m)} \matr{W}^{n-1} + \phi_{n-2}^{(m)}\matr{W}^{n-2} + \dots + \phi_{1}^{(m)} \matr{W} + \phi_{0}^{(m)} \matr{I}    
\end{equation}
implying that any power $m\geq n$ of $\matr{W}$ can be specified by $\matr{W}$ and scalars $(\phi_{n-1}^{(m)}, \dots \phi_{0}^{(m)})$. The apexes denote the fact that the $n$ coefficients are those proper of the $m$-th power for the $\phi_j^m$ coefficients.\change{
Note that, for $m < n$, we have $(\phi_{n-1}^{(i)}, \dots, \phi_{0}^{(i)}) =  (0, \dots, 0, 1, 0, \dots, 0)$, where the only non-zero term is the $m$-th one, i.e., 
      \begin{equation}\label{eqn:phi_delta}
        \phi^{(m)}_j = \delta_{mj} \quad\text{for}\quad m<n
    \end{equation}
}
Moreover, note that for $m = n$, $(\phi_{n-1}^{(m)}, \dots,  \phi_{0}^{(m)}) = (\varphi_{n-1}, \dots,  \varphi_{0})$. 

For each \change{$m \ge n$}, we can derive the scalars in recursive way by noting that:
\begin{align}\small
    &\matr{W}^{m+1} 
    =\matr{W}^{m} \matr{W} \\
    &=(\phi_{n-1}^{(m)} \matr{W}^{n-1} + \phi_{n-2}^{(m)}\matr{W}^{n-2} + \dots + \phi_{1}^{(m)} \matr{W} + \phi_{0}^{(m)} \matr{I}) \matr{W} \\
    &=\phi_{n-1}^{(m)} \matr{W}^{n} + \phi_{n-2}^{(m)}\matr{W}^{n-1} + \dots + \phi_{1}^{(m)} \matr{W}^{2} + \phi_{0}^{(m)} \matr{W} \\
    &=\phi_{n-1}^{(m)} \left(\varphi_{n-1} \matr{W}^{n-1} + \varphi_{n-2}\matr{W}^{n-2} + \dots
    + \varphi_{1} \matr{W} + \varphi_{0} \matr{I} \right) \nonumber\\
    &+ \phi_{n-2}^{(m)}\matr{W}^{n-1} + \dots + \phi_{1}^{(m)} \matr{W}^{2} + \phi_{0}^{(m)} \matr{W} \\
    &= \underbrace{(\varphi_{n-1} \phi_{n-1}^{(m)} + \phi_{n-2}^{(m)})}_{\phi_{n-1}^{(m+1)}}\matr{W}^{n-1} + 
    \underbrace{(\varphi_{n-2} \phi_{n-1}^{(m)} + \phi_{n-3}^{(m)})}_{\phi_{n-2}^{(m+1)}} \matr{W}^{n-2}\nonumber\\ 
    &+ \dots 
    + \underbrace{(\varphi_{1} \phi_{n-1}^{(m)} + \phi_{0}^{(m)})}_{\phi_{1}^{(m+1)}} \matr{W} 
    + \underbrace{(\varphi_{0} \phi_{n-1}^{(m)})}_{\phi_{0}^{(m+1)}} \matr{I}   
\end{align}{}
which implies
\begin{equation} \label{eqn:phis}
    \begin{cases}
    \phi_{0}^{(m+1)}&= \varphi_{0} \phi_{n-1}^{(m)}
    \\\phi_{1}^{(m+1)} &=\varphi_{1} \phi_{n-1}^{(m)} + \phi_{0}^{(m)})   
    \\&\dots 
    \\\phi_{n-2}^{(m+1)} &=\varphi_{n-2} \phi_{n-1}^{(m)} + \phi_{n-3}^{(m)}
    \\\phi_{n-1}^{(m+1)}&= \varphi_{n-1} \phi_{n-1}^{(m)} + \phi_{n-2}^{(m)}
    \end{cases}
\end{equation}{}
Eq. \ref{eqn:phis} can be thought as a linear system:
\begin{equation}
    \begin{bmatrix}
        &   \phi^{(m+1)}_{0}\\
        &    \phi^{(m+1)}_{1}\\
        & \vdots \\
        &   \phi^{(m+1)}_{n-2}\\
        &   \phi^{(m+1)}_{n-1}\\    
    \end{bmatrix}
    =
    \matr{M}
    \begin{bmatrix}
    &   \phi^{(m)}_{0}\\
    &    \phi^{(m)}_{1}\\
    & \vdots \\
    &   \phi^{(m)}_{n-2}\\
    &   \phi^{(m)}_{n-1}\\
    \end{bmatrix}
\end{equation}
where $\matr{M}$ is defined as:
\begin{equation}
\matr{M} =
\begin{bmatrix}
0  & \hdotsfor{2}   & 0  &\varphi_{0}\\
1 & 0 & \dots      & 0 & \varphi_{1}  \\
\hdotsfor{5}\\
0  & \dots & 1  & 0 & \varphi_{n-2} \\    
0 & 0 & \dots  & 1  &\varphi_{n-1} \\
\end{bmatrix}
\end{equation}

Note that the characteristic polynomial of $\matr{M}$ is equal to the one of $\matr{W}$, so that they also share the same eigenvalues. In fact, $\matr{M}$ is also know as the Frobenius companion matrix of $\matr{W}$.

\section{The network encoded input}\label{app:input}

The possibility for the readout to produce the  \change{correct output for the task at hand} depends on two distinct elements: the controllability matrix $\matr{\mathcal{C}}$ (which depends on $\matr{W}$ and $\vec{w}$) and the $\vec{s}$ vector (which depends on both $\matr{W}$ and the signal $\vec{u}$).
Here we show how $\vec{s}$ is obtained from $\vec{u}$.

Under the assumption of bounded inputs $u_{-k} \in [-U,U], \forall k$, we see that 
\[
|s_j | = \left|\sum_{k=0}^\infty \phi_j^{(k)}u_{-k}\right| \le U \sum_{k=0}^\infty \left|\phi_j^{(k)}\right| 
\]
allowing us to focus on the properties of the $\phi_j^{(k)}$.

These terms are the element of $\vec{s}$, which we  rewrite as:
\begin{equation}
    \begin{bmatrix}{}
    &s_0 \\
    &s_1 \\
    & \dots \\
    &s_{n-2}\\
    &s_{n-1}
    \end{bmatrix}
    =
    \begin{bmatrix}
    &\sum_{k=0}^\infty \phi_0^{(k)}u_{-k}   \\
    &\sum_{k=0}^\infty \phi_1^{(k)}u_{-k} \\
    & \dots \\
    &\sum_{k=0}^\infty \phi_{n-2}^{(k)}u_{-k} \\
    &\sum_{k=0}^\infty \phi_{n-1}^{(k)}u_{-k}\\
    \end{bmatrix}
\end{equation}{}

In Appendix~\ref{app:CHT} we show that \change{for $k<n$, $\phi^{(k)}_j = \delta_{kj}$} (Eq.~\ref{eqn:phi_delta}).
This implies that the first $n-1$ time steps are simply the inputs:
\begin{equation}
\begin{split}
    \begin{bmatrix}{}
    &s_0 \\
    &s_1 \\
    & \dots \\
    &s_{n-2}\\
    &s_{n-1}
    \end{bmatrix}
    &=
    \begin{bmatrix}
    &u_{0} + \sum_{k=n}^\infty \phi_0^{(k)}u_{-k}   \\
    &u_{-1} +\sum_{k=n}^\infty \phi_1^{(k)}u_{-k} \\
    & \dots \\
    &u_{-(n-2)} +\sum_{k=n}^\infty \phi_{n-2}^{(k)}u_{-k} \\
    &u_{-(n-1)} +\sum_{k=n}^\infty \phi_{n-1}^{(k)}u_{-k}\\
    \end{bmatrix}\\
    &=
    \begin{bmatrix}
    &u_{0}    \\
    &u_{-1}\\
    & \dots \\
    &u_{-(n-2)} \\
    &u_{-(n-1)}\\
    \end{bmatrix}
    +
    \begin{bmatrix}
    & \sum_{k=n}^\infty \phi_0^{(k)}u_{-k}   \\
    &\sum_{k=n}^\infty \phi_1^{(k)}u_{-k} \\
    & \dots \\
    & \sum_{k=n}^\infty \phi_{n-2}^{(k)}u_{-k} \\
    &\sum_{k=n}^\infty \phi_{n-1}^{(k)}u_{-k}\\
    \end{bmatrix}
\end{split}
\end{equation}
Then, we observe that the terms corresponding to time-step $k=n$ follow from Eq. \ref{eq:an-linear-comb}:
\change{
\begin{equation}
\begin{split}
     \begin{bmatrix}{}
    &s_0 \\
    &s_1 \\
    & \dots \\
    &s_{n-2}\\
    &s_{n-1}
    \end{bmatrix}
        &=
    \begin{bmatrix}
    &u_{0}  \\
    &u_{-1} \\
    &\dots \\
    &u_{-(n-2)}\\
    &u_{-(n-1)}  \\
    \end{bmatrix}
    +
    \begin{bmatrix}
   &u_{-n} \varphi_0      \\
    &u_{-n} \varphi_1 \\
    &\dots \\
    &u_{-n} \varphi_{n-2} \\
    &u_{-n} \varphi_{n-1} \\
    \end{bmatrix}
    +\\
    &+
    \begin{bmatrix}
    & \sum_{k=n+1}^\infty \phi_0^{(k)}u_{-k}   \\
    &\sum_{k=n+1}^\infty \phi_1^{(k)}u_{-k} \\
    & \dots \\
    & \sum_{k=n+1}^\infty \phi_{n-2}^{(k)}u_{-k} \\
    & \sum_{k=n+1}^\infty \phi_{n-1}^{(k)}u_{-k} \\
    \end{bmatrix}
\end{split}
\end{equation}
}
successive terms corresponding to time steps $k>n$ can be computed by using \eqref{eqn:phis}. This procedure shows that, in general, the inputs from $0$ to $n-1$ time steps in the past will \emph{always} appear in their original form, and the ``mixing'' will begin starting from the $n$-th time step in the past.

\section{Delay line}\label{app:delay_line}

It is easy to see that, applying $\matr{W}_{\text{d}}$ to a vector $\vec{v} = (v_1,v_2,\dots,v_n)$ results in a vector 
\[
\vec{v}' := \matr{W}_{\text{d}ij}\vec{v} = (0,v_1,\dots,v_{n-1})
\]
and because of the associativity of the matrix product, we see that applying $\matr{W}_{\text{d}ij}$ to a vector $k$ times results in permuting the vector $k$ times and the substituting the first $k$ elements with the same number of $0$s.
So, the controllability matrix for the delay line is:

\begin{equation}
      \matr{\mathcal{C}}_\text{d} = [\vec{w}_{\text{d}}  \quad \matr{W}_{\text{d}} \vec{w}_{\text{d}} \quad \dots \quad  \matr{W}_{\text{d}}^{n-1} \vec{w}_{\text{d}}  ]
\end{equation}{}

which would be  a lower diagonal matrix for a generic $\vec{v}$ but for $\vec{w}_{\text{d}} = (1,0,\dots,0)$ is just the identity.

Now, consider the fact that 

\begin{equation}
    \matr{W}_\text{d}^n = \matr{0} 
\end{equation}

The \ac{CH} theorem implies that any higher power will be \change{null} as well. 
So we simply have:
\[
s_0 = u_0
\]
\[
s_1 = u_1
\]

and so on, because all the $\phi_j^{(m)}$ for $m>n$ are null. 
If we define $\vec{s}_{\text{d}} := (u_0, u_{-1} , u_{-2}, \dots, u_{-(n-1)})$:
\begin{equation}
    \vec{y}_0 =  \vec{r} \cdot \matr{\mathcal{C}}_\text{d} \cdot \vec{s}_\text{d} = \vec{r} \cdot \matr{I} \cdot \vec{s}_\text{d} = \sum_{i=0}^{n-1} r_i u_{-i}
\end{equation}{}
which, as expected, is simply a regressive model of order $n$.

\section{Cyclic reservoirs}\label{app:cyclic}

The characteristic polynomial of $\matr{W}_\text{c}$ is  $\lambda^n = 1$ so that the \ac{CH} Theorem implies: 
\begin{equation}
\matr{W}^n_\text{c} = \matr{I}
\end{equation}

Meaning that, for all $m>n$,
\begin{equation}
    \matr{W}_\text{c}^m = \sum_{j=0}^{n-1} \phi_j^{(m)} \matr{W}_\text{c}^j = \matr{W}_\text{c}^{\mu}
\end{equation}
where $\mu := m\mod{n}$. Note that, in general:
\begin{equation}
    (a\matr{W}_\text{c})^m = a^m \matr{W}_\text{c}^{\mu}
\end{equation}

So, if in our reservoir we fix $\matr{W} = \rho\matr{W}_\text{c}$ (where $\rho$ is a parameter controlling the spectral radius) we obtain a number of simplifications.
First of all, the elements of $\vec{s}$ assume a regular form.
For example:
\begin{align*}
    s_0 &= u_0 + \rho^n u_{-n} + \rho^{2n} u_{-2n} + \dots  \\
    s_1 &= u_{-1} + \rho^{n} u_{-(n+1)} + \rho^{2n} u_{-(2n+1)} + \dots  
\end{align*}
so that their general form is
\begin{equation} \label{eqn:s_cyclic}
    s_j = \sum_{k=0}^\infty \phi_j^{(k)}u_{-k}  = \sum_{p=0}^{\infty} \rho^{pn} u_{-{j+pn}}
\end{equation}{}

Moreover, the controllability matrix $ \matr{\mathcal{C}}_\text{c}$ assumes a simple form. If we define the $i$-time permuted input weight vector as:
\begin{equation}
    \vec{w^{(i)}} := \matr{W}_\text{c}^i w
\end{equation}{}
we obtain:
\begin{equation} \label{eqn:cyclic_controllability}
    \matr{\mathcal{C}}_\text{c} = [\vec{w} \quad \rho\vec{w}^{(1)} \quad \rho^2\vec{w}^{(2)} \quad \dots \quad  \rho^{n-1}\vec{w}^{(n-1)}  ]
\end{equation}
so that:
 \begin{align}
y_0 &= (r_{0},r_{1},\cdots, r_{n-1})
          \matr{\mathcal{C}}_\text{c}
            \begin{pmatrix}    
          s_{0} \\
          s_{1} \\
          s_{2} \\
          \vdots \\
          s_{n-1}
         \end{pmatrix}
 \end{align}

The output can be written in compact form by defining:
\begin{equation}
    \Tilde{s}_j = \sum_{p=0}^{\infty} \rho^{j+pn} u_{-{j+pn}}
\end{equation}{}
\begin{equation}
    \Tilde{\matr{\mathcal{C}}}_\text{c} = [\vec{w} \quad \vec{w}^{(1)} \quad \vec{w}^{(2)} \quad \dots \vec{w}^{(n-1)}  ]
\end{equation}{}
so that, finally:
\begin{equation}
    y_0 = \vec{r} \Tilde{\matr{\mathcal{C}}}_\text{c}  \Tilde{\vec{s}}
\end{equation}

\section*{Acknowledgment}
LL gratefully acknowledges partial support of the Canada Research Chairs program.




\bibliographystyle{IEEEtran}
%

\bibliography{references}




%





\end{document}